\documentclass[letterpaper]{article} 
\usepackage{aaai25}  
\usepackage{times}  
\usepackage{helvet}  
\usepackage{courier}  
\usepackage[hyphens]{url}  
\usepackage{graphicx} 
\usepackage{natbib}  
\usepackage{caption} 
\usepackage{algorithm}
\usepackage{algorithmic}

\usepackage{newfloat}
\usepackage{listings}
\DeclareCaptionStyle{ruled}{labelfont=normalfont,labelsep=colon,strut=off} 
\floatstyle{ruled}
\newfloat{listing}{tb}{lst}{}
\floatname{listing}{Listing}
\usepackage{graphicx}
\usepackage{booktabs}
\usepackage{array}

\usepackage{pgfplots}
\usepackage{enumitem}
\usepackage{multirow}
\usepackage{amsmath}
\DeclareMathOperator*{\argmin}{arg\,min}
\usepackage{tikz}
\usepackage{pgfplots}
\usepackage{lipsum,booktabs}

\usepackage{eccvabbrv}

\title{Coordinate Descent for Network Linearization}

\author{
    Vlad Rakhlin\equalcontrib,
    Amir Jevnisek\equalcontrib,
    Shai Avidan
}
\affiliations{
    School of Electrical Engineering\\
    Tel-Aviv University \\
    Tel-Aviv, Israel \\

    vladislavr@mail.tau.ac.il, amirjevn@mail.tau.ac.il, avidan@eng.tau.ac.il
}

\begin{document}

\maketitle

\begin{abstract}
  ReLU activations are the main bottleneck in Private Inference that is based on ResNet networks. This is because they incur significant inference latency. Reducing ReLU count is a discrete optimization problem, and there are two common ways to approach it. Most current state-of-the-art methods are based on a smooth approximation that jointly optimizes network accuracy and ReLU budget at once. However, the last hard thresholding step of the optimization usually introduces a large performance loss. We take an alternative approach that works directly in the discrete domain by leveraging Coordinate Descent as our optimization framework. In contrast to previous methods, this yields a sparse solution by design. We demonstrate, through extensive experiments, that our method is State of the Art on common benchmarks.

\end{abstract}

\section{Introduction}

Private Inference (PI) is an emerging scheme that lets Service Providers offer privacy preserving Machine Learning as a Service to their users. In PI setting, the service provider runs inference on user data without learning anything about the content of the data. Similarly, the user learns nothing about the weights of the network.

PI applies cryptographic primitives to the different components of a neural network. As it turns out, linear (\eg, convolutions) and non-linear (\eg, ReLU) operations use different primitives with different representations, which requires constant translation between the two. Moreover, the large number of ReLUs in a neural network  requires a large number of translations and, to top it off, the execution of a single ReLU is quite expensive in terms of communication bandwidth~\cite{srinivasan2019delphi}. This makes it necessary to reduce the number of ReLU operations within a neural network, in order to make PI efficient and practical.

\begin{figure*}[!t]
    \centering
    \resizebox{1.0\textwidth}{!}{
    \begin{tabular}{c c c} 
         \resizebox{0.5\linewidth}{!}{\begin{tikzpicture}
    \begin{semilogxaxis}[legend pos=south east,
        legend style={nodes={scale=0.62, transform shape}},
        width=0.7\columnwidth,
        height=5.7cm,
        xlabel=ReLU Count (\#K),
        ylabel=Test Accuracy (\%),
        xlabel style={at={(0.5, 0.1)}},
        ylabel style={at={(-0.05, 0.5)}},
        title=CIFAR-10,
        xmin=4, xmax=550,
        ymin=85,ymax=96,
        ylabel near ticks,
        xlabel near ticks,
        axis background/.style={fill=blue!0},
        grid=both,
        log basis x = 2,
        /pgf/number format/1000 sep={\,},
        log ticks with fixed point,
        grid style={line width=.1pt, draw=gray!10},
        major grid style={line width=.2pt,draw=gray!50},
        ]
    \addplot[dashed, thick] plot coordinates {(2,95.22) (512,95.22)};
    \addlegendentry{Baseline}
     \addplot[mark=*,mark size=2.5pt,blue, thick]
        plot coordinates {
            (6, 85.65)
            (9, 87.49)
            (14.5, 89.65)
            (100, 93.78)
            (150, 94.94)
        };
    \addlegendentry{\textbf{Our}}

    \addplot[mark=x,mark size=2pt,red, only marks]
        plot coordinates {
            (6, 84.39)
            (9, 86.97)
            (12.9, 88.23)
            (14.9, 88.43)
            (25.0, 90.88)
            (29.8, 90.92)
            (40.0, 91.68)
            (49.4, 92.27)
            (60.0, 92.63)
            (69.8, 93.02)
            (79.1, 93.16)
            (99.9, 93.50)
            (150.0, 94.26)
            (180.0, 94.78)
            (300, 95.06)
            (400, 95.07)
            (500, 95.21)
        };
    \addlegendentry{SNL}

    \addplot[mark=diamond,mark size=2pt,orange, thick, only marks]
        plot coordinates {
       (36.0 , 88.5)
            (70.0, 90.0)
            (80.0, 90.5)
            (114.0, 92.7)
            (147.0, 93.16)
            (221.48,94.07)
        };
    \addlegendentry{DeepReduce}

    \addplot[mark=pentagon, mark size=2pt, brown, thick, only marks] 
        plot coordinates {
            (50, 90.0)
            (86, 91.5)
            (100, 92.2)
            (334, 94)
            (500, 94.8)
        };
    \addlegendentry{CryptoNAS}
    
    \end{semilogxaxis}
\end{tikzpicture}} &
         \resizebox{0.5\linewidth}{!}{\begin{tikzpicture}
    \begin{semilogxaxis}[legend pos=south east,
        legend style={nodes={scale=0.62, transform shape}},
        width=0.7\columnwidth,
        height=5.7cm,
        xlabel=ReLU Count (\#K),
        ylabel=Test Accuracy (\%),
        xlabel style={at={(0.5, 0.1)}},
        ylabel style={at={(-0.05, 0.5)}},
        title=CIFAR-100,
        xmin=4, xmax=550,
        ymin=57,ymax=80,
        ylabel near ticks,
        xlabel near ticks,
        axis background/.style={fill=blue!0},
        grid=both,
        log basis x = 2,
        /pgf/number format/1000 sep={\,},
        log ticks with fixed point,
        grid style={line width=.1pt, draw=gray!10},
        major grid style={line width=.2pt,draw=gray!50},
        ]
    \addplot[dashed, thick] plot coordinates {(2,76.04) (512,76.04)};
    \addlegendentry{Baseline}
     \addplot[mark=*,mark size=2.5pt,blue, thick]
        plot coordinates {
            (6, 61.78)
    (9, 64.8)
    (12.9, 68.24)
    (14.6, 68.74)
    (15, 68.85)
    (16.8, 69.57)
    (20, 69.7)
    (100, 76.5)
    (150, 77.27)
        };
    \addlegendentry{\textbf{Our}}

        

    \addplot[mark=x,mark size=2pt,red, only marks]
        plot coordinates {
            (6, 57.65)
            (7, 59.5)
            (8, 60.48)
            (9, 62.8)
            (10, 64.16)
            (11, 64.8)
            (12, 65.57)
            (13, 66.02)
            (15, 67.17)
            (16, 67.94)
            (17, 68.54)
            (20, 69.38)
            (25.0, 70.23)
            (30, 70.89)
            (50, 73.22)
            (100, 75.42)
            (150, 76.13)
        };
    \addlegendentry{SNL}

    \addplot[mark=diamond,mark size=2pt,orange, thick, only marks]
        plot coordinates {
           (12.3, 64.97)
            (28.7 , 68.68)
            (49.2, 69.50)
            (57.34, 72.68)
            (114.0, 74.72)
            (197.0, 75.51)
            (229.38,76.22)
        };
    \addlegendentry{DeepReduce}

    \addplot[mark=otimes, mark size=2pt, violet, thick, only marks]
        plot coordinates {
        (25.6, 66.13)
            (30.2, 67.37)
            (41.0, 68.23)
            (51.2, 69.57)
            (71.7, 71.06)
            (102.4, 72.90)
        };
    \addlegendentry{SPHYNX}

    \addplot[mark=pentagon, mark size=2pt, brown, thick, only marks]
        plot coordinates {
            (50, 63.5)
            (86, 66)
            (100, 6.5)
        };
    \addlegendentry{CryptoNAS}

    \end{semilogxaxis}
\end{tikzpicture}} &
         \resizebox{0.5\linewidth}{!}{\begin{tikzpicture}
    \begin{semilogxaxis}[legend pos=south east,
        legend style={nodes={scale=0.62, transform shape}},
        width=0.7\columnwidth,
        height=5.7cm,
        xlabel=ReLU Count (\#K),
        ylabel=Test Accuracy (\%),
        xlabel style={at={(0.5, 0.1)}},
        ylabel style={at={(-0.05, 0.5)}},
        title=TinyImageNet,
        xmin=31, xmax=1027,
        ymin=45,ymax=66,
        ylabel near ticks,
        xlabel near ticks,
        axis background/.style={fill=blue!0},
        grid=both,
        log basis x = 2,
        /pgf/number format/1000 sep={\,},
        log ticks with fixed point,
        grid style={line width=.1pt, draw=gray!10},
        major grid style={line width=.2pt,draw=gray!50},
        ]
    \addplot[dashed, thick] plot coordinates {(2,65.05) (1024,65.05)};
    \addlegendentry{Baseline}
     \addplot[mark=*,mark size=2.5pt,blue, thick]
        plot coordinates {
            (59.1, 56.6)
    (99.6, 59.77)
    (150, 62.98)
    (200, 63.97)
        };
    \addlegendentry{\textbf{Our}}

    \addplot[mark=x,mark size=2pt,red, only marks]
        plot coordinates {
            (59.1, 54.24)
    (99.6, 58.94)
    (150, 62.12)
    (200, 63.39)
    (300, 64.04)
    (400, 63.83)
    (500, 64.42)
        };
    \addlegendentry{SNL}

    \addplot[mark=diamond,mark size=2pt,orange, thick, only marks]
        plot coordinates {
           (57.35, 53.75)
             (98.3, 55.67)
             (114.69, 56.18)
             (200.0, 57.51)
             (230.0, 59.18)
             (400.0, 61.65)
             (459.0, 62.26)
             (917.0, 64.66)
        };
    \addlegendentry{DeepReduce}

    \addplot[mark=otimes, mark size=2pt, violet, thick, only marks]
        plot coordinates {
        (102.4, 48.44)
            (204.8, 53.51)
            (286.7, 56.72)
            (491.5, 59.12)
            (614.4, 60.76)
        };
    \addlegendentry{SPHYNX}

    \end{semilogxaxis}
\end{tikzpicture}} \\
         
    \end{tabular}
    }
    \caption{\textbf{Accuracy vs ReLU Budget for a ResNet18 Network: } ResNet18 Accuracy [\%] for classifying CIFAR-10, CIFAR-100 and TinyImageNet for different ReLU budgets [\# ReLUs]. Our method achieves the best performance for every ReLU budget. For some budgets our method surpasses the baseline which is the original network containing 100\% of the ReLUs. }
    \label{fig:acc_vs_relu_budget_three_datasets}
\end{figure*}
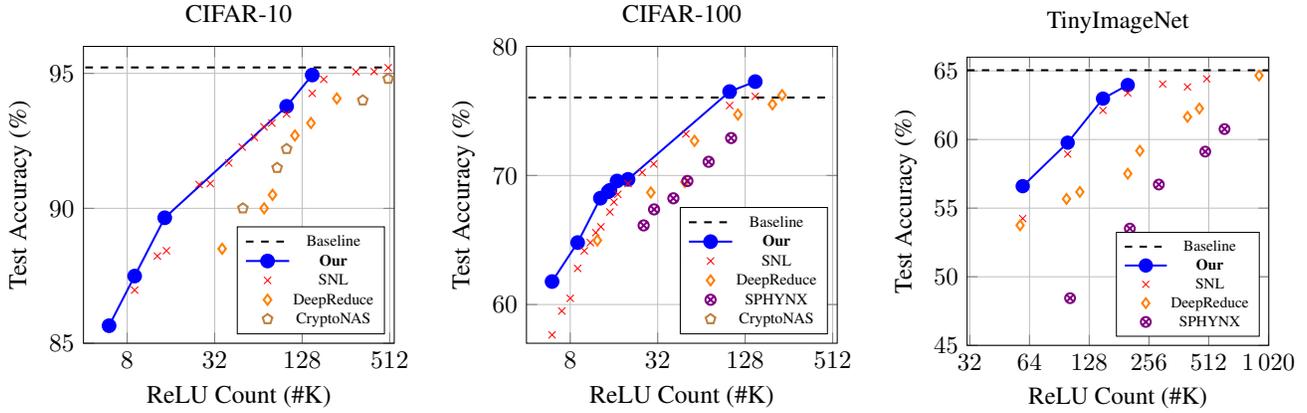

Recent work in this space learn which ReLUs to keep and which ones to remove~\cite{cho2022selective, peng2023autorep}, while preserving network performance. Specifically, this is done by constructing a mask for all ReLU units in the network and searching for a sparse mask that still preserves network accuracy performance. We term these approaches Selective approaches. When ReLUs are replaced with identity functions, this is addressed as Network Linearization~\cite{cho2022selective}.

Selective approaches often use a LASSO-like regression~\cite{tibshirani1996regression} to solve the problem. That is, instead of having an $L_0$ constraint on the number of ReLU units left, they use a relaxed $L_1$ constraint. And instead of strictly enforcing the constraint they soft enforce it using a Lagrange multiplier~\cite{cho2022selective, peng2023autorep}. 

LASSO based regression encourages sparse solutions but does not guarantee it. Specifically, the differentiable nature of the optimization "leaks" information during the process, because the mask values are not binary (\ie, 0 and 1) but rather real values in the range $[0,1]$. As a result, the hard thresholding stage at the end of optimization hurts accuracy considerably. 

Our method, on the other hand, guarantees a sparse solution. We show that this design-choice results in superior performance compared with the previously suggested Selective approaches.  Specifically, we use Block Coordinate Descent. Our algorithm starts with some network that we wish to reduce its ReLU count. Then, at each stage, we randomly sample a subset of ReLUs to be removed, and evaluate their impact (or lack thereof) on accuracy. We repeat this process several times and remove the subset with the least amount of impact on performance. We continue with this process and every time the accuracy drops below a predefined threshold, we fine-tune the network.

The proposed approach has a couple of benefits. First, it is flexible and can work with different network architectures. Second, it works directly with the discrete mask values and does not require a hard thresholding post-processing step. 

We conducted extensive experiments and found that our method achieves SOTA results on the standard benchmarks. Specifically, we demonstrate that our method can be applied on top of and \emph{improve} existing SOTA methods such as SNL~\cite{cho2022selective} and AutoRep~\cite{peng2023autorep}. Figure~\ref{fig:acc_vs_relu_budget_three_datasets} shows our performance for the case of ResNet18 backbone on three datasets (CIFAR-10~\cite{krizhevsky2009learning}, CIFAR-100~\cite{krizhevsky2009learning} and TinyImageNet~\cite{le2015tiny}). This figure highlights that our method performs especially well in the low ReLU budget regime. To conclude, our contributions are threefold:
\begin{itemize}
    \item We suggest an optimization method which naturally produces a sparse solution to a sparse optimization problem, without relaxations and approximations to the loss functions or regularization terms.
    \item We present SOTA results for reducing ReLU budgets on three datasets: CIFAR-10~\cite{krizhevsky2009learning}, CIFAR-100~\cite{krizhevsky2009learning} and TinyImageNet~\cite{le2015tiny} and with respect to two backbones: ResNet18~\cite{he2016deep} and WideResNet22-8~\cite{zagoruyko2016wide}. Our performance improves over the previous SOTA by up to 4.13 \% (in the case of ResNet18, CIFAR-100 and ReLU budget of $6K$).
    \item We demonstrate that when our method is applied on top of existing SOTA methods such as SNL and AutoRep. In the case of AutoRep, we achieve the same accuracy with just \emph{half} of the ReLU budget.
\end{itemize}

\section{Related Work}

\begin{figure}[h]
    \centering
    \resizebox{0.5\textwidth}{!}{\input{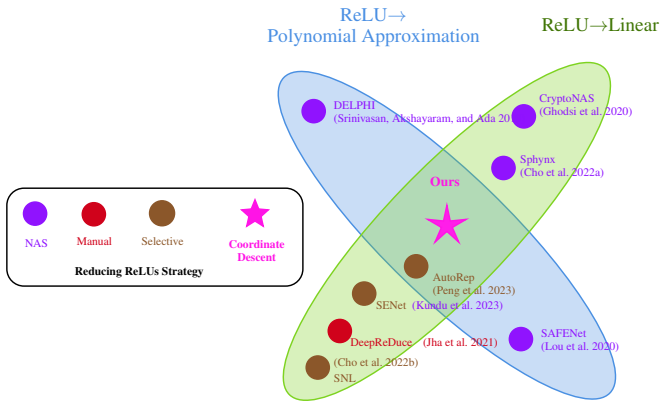} }
    \caption{\textbf{ReLU Reduction Methods: } The green ellipse denotes methods that replace ReLUs with identity functions (known as {\em Network Linearization}). The blue ellipse denotes methods that replace ReLUs with polynomial approximations (which are faster to compute than ReLUs in a PI setting). The color of the circles denotes the optimization approach used. Selective methods (brown circles) jointly optimize Cross-Entropy loss and a regularization term that takes care of adhering to a budget. NAS (purple circles) use Network Architecture Search for the optimization, while Manual (red circle) outlines a manual procedure for ReLU reduction. We use Block Coordinate Descent (star) that provides a sparse solution by design.}
    \label{fig:relu_reduction_methods}
\end{figure}

In this section we first describe cryptographic protocols aimed at enabling Private Inference. Then, we switch to describing methods that are aimed at ReLUs-reduction which are the main bottleneck in terms of communication bandwidth and inference time.

\subsection{Cryptographic Protocols for Private Inference}

There is an assortment of prior works that deal with building the underlying cryptographic systems required for Private Inference. These cryptographic systems may be build upon different approaches; these approaches may be based on Homomorphic Encryption~\cite{gentry2009fully}, Oblivious Transfer~\cite{cryptoeprint:2005/187}, Secret Sharing~\cite{shamir1979ss} or some combination of the aforementioned concepts.

CryptoNets~\cite{bachrach2016cryptonets} were the first ones that performed predictions on neural networks directly on encrypted data using leveled Homomorphic Encryption.
GAZELLE~\cite{juvekar2018gazelle} also used Homomorphic Encryption but they added significant optimizations for basic homomorphic operations, as well as optimized conversion routines between homomorphic and Garbled Circuits~\cite{yao1982gc} forms for inference.
SecureML~\cite{mohassel2017secureml} proposed a Multi Party Computation (MPC) protocol based on Secret Sharing in conjuction with Garbled Circuits~\cite{yao1982gc} while also utilizing Homomorphic Encryption and Oblivious Transfer to implement the pre-sharing phase.
SecureNN~\cite{wagh2018securenn} and FALCON~\cite{wagh2020falcon} proposed a set of protocols for 3-party MPC that utilize Secret Sharing exclusively.

\subsection{Reducing ReLU Count} ReLUs are the main bottleneck for private inference for ResNet architectures~\cite{srinivasan2019delphi}. When one reduces ReLUs there are two design choices. The first is to devise a strategy that determines which ReLUs to reduce. The second design choice is which function is used instead of the ReLU. We divide the landscape of prior art according to the two design choices as depicted in Figure~\ref{fig:relu_reduction_methods}. The ReLU choice strategies are denoted with different markers. The ReLU replacement functions are split between the ellipse areas.

Prior work tackles the challenge of reducing ReLU counts by designing ReLU-efficient networks using Neural Architecture Search (NAS), such methods include CryptoNAS~\cite{ghodsi2020cryptonas} and SPHYNX~\cite{cho2022sphynx}. Other NAS-based methods like DELPHI~\cite{srinivasan2019delphi} and SAFENET~\cite{lou2020safenet} propose replacing ReLUs with polynomial approximations. 

Other methods attempt to eliminate ReLUs directly. DeepReDuce~\cite{jha2021deepreduce} characterized the behavior of ReLUs in ResNet networks and pointed out some insights regarding ReLU importance at different network layers and reduction policies that lead to substantial reduction in ReLUs in exchange for reasonable accuracy drops. Based on that, they proposed a set of actions to eliminate ReLUs at varying granularities and arrive at relatively sparse models, similarly to our approach.

Another line of work is Selective approaches. The first seminal work here is Selective Network Linearization (SNL)~\cite{cho2022selective}. SNL introduced the selective approach in this context, integrating a learnable parameter for each ReLU in the network. This parameter indicates whether a ReLU at a certain pixel location is present or otherwise replaced with an identity function. The network is trained with Cross Entropy loss and a loss term which encourages small values for the ReLU. The budget at each training epoch is evaluated through thresholding the mask values with respect to some threshold hyperparameter. The network is trained with both terms until it arrives at the target ReLU budget. At this point, the ReLU masks are hard thresholded and some performance degradation is introduced, since ReLU signal is still "leaked" from ReLUs with mask values under the threshold. To compensate for this degradation, the weights of the neural network are fine-tuned, with the binarized ReLU masks.

SENet~\cite{kundu2023senet} showed that by utilizing weight-pruning approaches, it is possible to derive a metric that describes the importance of ReLU layers within a neural network, which they call ReLU sensitivity. Based on that, they present an algorithm that allocates the desired ReLU budget among all layers. Then, for each ReLU layer, an auxiliary binary search mask of the same shape as the ReLU layer is maintained; through a Knowledge Distillation~\cite{hinton2015distilling} training routine, this mask is modified such that ReLUs are only left at pixel locations where the response is the farthest from the response of the full-ReLU model at the same location. Finally, the model is finetuned using Knowledge Distillation with an additional loss term called post-ReLU activation mismatch (PRAM), which helps the ReLU-reduced model output activation maps similar to the full-ReLU model.

Automatic ReLU Reduction~\cite{peng2023autorep} (AutoReP) introduced a concept that somewhat resembles SNL, but instead of eliminating selected ReLUs completely, they are replaced with polynomial approximations whose coefficients are learnable and they also depend on the distribution of values at the input of the respective ReLU element. The ReLUs that are chosen to be replaced are selected by an optimized method consisting of a trainable indicator function and a hysteresis loop, which acts as a stabilizer for the indicator function.

\section{Method}
\paragraph{Goal} The goal of this work is to train a Neural Network that achieves the best task performance under a ReLU budget constraint. We denote a Neural Network with parameters $\theta$ and a ReLU mask $m$ as: $f_{\theta, m }(\cdot)$. A ReLU mask $m$ is a binary mask. At a neuron location, when the mask is equal to 1, it denotes that there is a ReLU at that location in the network. When set to 0, an alternative computation is performed instead. In this work we focus on a classification task. Formally stated, for a ReLU budget $B$, we are interested in finding the parameters $\theta$ and ReLU mask $m$ that satisfy the following optimization problem:

\begin{equation}
    \begin{aligned}
    \min_{\theta, m} \quad & \mathcal{L}_{\text{cross-entropy}}(f_{\theta, m }(x), y)\\
    \textrm{s.t.} \quad & ||m ||_0 \leq B\\
    \end{aligned}
    \label{eq:general_case_subject_to}
\end{equation}

\paragraph{Our method} The key insight of our method is that Selective approaches do not yield a sparse solution by design. Instead, we use Block Coordinate Descent which yields a sparse solution with provable runtime-performance guarantees that we demonstrate to outperform the current SOTA. See Algorithm~\ref{alg:our_alg}.

Our algorithm is iterative and involves three parameters. The first is a parameter that controls the number of ReLUs we reduce in each Coordinate Descent iteration. We term this parameter Delta ReLU Count (DRC). Since exploring all DRC configurations is intractable, we limit the algorithm search space to Random Trials (RT) different hypotheses of DRC groups. Reducing ReLUs incurs accuracy degradation, therefore we finetune the network to regain performance. We finetune for a predetermined number of epochs, when the accuracy drops by Accuracy Degradation Tolerance (ADT) \%. 

\begin{algorithm}
\caption{\textsc{Block Coordinate Descent}} 
\label{alg:our_alg}
    \begin{algorithmic}
        \STATE \textbf{Inputs:} A network $f_{\theta, m}(\cdot)$ with parameters $\theta$ and a ReLUs mask $m$, \newline
        a train dataset $\mathcal{D}_{train}$,  \newline
        $B_\textrm{target}[\# \textrm{ReLUs}]$: target ReLU budget,   \newline
        $RT$ the number of random tries, \newline
        $DRC [\# \textrm{ReLUs}]$ is the Delta ReLU Count and \newline
        $ADT$ [\%] is the Accuracy Degradation Tolerance. 
        \STATE \textbf{Output:} A network $f_{\theta, m}(\cdot)$ with $||m||_0=B_{\textrm{target}}$
        \STATE \textbf{{\textsc{Algorithm:}}}
        \STATE We start from a network with $||m||_0 = B_\textrm{ref}[\# \textrm{ReLUs}]$ ReLUs.
        \FOR{ $t = 1, ..., T = \lceil\frac{B_\textrm{ref} - B_\textrm{target}}{DRC}\rceil$}
            \REPEAT
                \STATE {Sample $DRC$ ReLUs and mask them out.} 
                \STATE {Evaluate Network accuracy on $\mathcal{D}_{train}$.}
            \UNTIL $RT \text{ times } \lor$ accuracy drops by less than $ADT$
            \STATE {Choose the mask corresponding to the lowest train accuracy degradation scanned and finetune}
        \ENDFOR
        
    \end{algorithmic}
\end{algorithm}

If the number of reduce attempts was exhausted, then the ReLU set that incurred the lowest train loss is selected. This procedure is executed until we arrive at the desired ReLU budget. A detailed pseudo code is attached to the supplementary section. To speed up training, we use SNL~\cite{cho2022selective} or AutoReP~\cite{peng2023autorep} with a medium ReLU-budget-model as a starting point for our optimization.

\section{Experiments}

\subsection{Experiments Settings}
We evaluate our method on ResNet18~\cite{he2016deep} and Wide-ResNet-22-8~\cite{zagoruyko2016wide} backbones with several datasets - CIFAR-10, CIFAR-100 and TinyImageNet.  
\paragraph{Datasets}
CIFAR-10 is a 10 classes classification dataset which comprises of $50K$ train images and $10K$ test images. For each class there are $5K$ training images and $1K$ test images. Each image is an RGB image of size $32 \times 32$. CIFAR-100 is a dataset of 100 classes with 500 training images and 100 test images per class with the same image size as CIFAR-10. We extend our evaluation to $64 \times 64$ size images, utilizing the TinyImageNet dataset. TinyImageNet is a classification dataset which aims at classifying 200 different classes with 500 train images and 50 test images per class.

\begin{table}[]
\centering

\begin{tabular}{|c||c|c|}
\hline
\textbf{Image Shape} & \textbf{ResNet18} & \textbf{Wide-ResNet-22-8} \\ \hline \hline
$32 \times 32$               & 570K              & 1359K                     \\ \hline
$64 \times 64$               & 1966K             & 5439K                     \\ \hline
\end{tabular}

\caption{\textbf{Overall Number of ReLUs [\#K]} For each Network and Image-Size configuration we state the overall number of ReLUs. CIFAR-10 and CIFAR-100 image shapes are $32\times 32$ whereas TinyImageNet's image shape is $64 \times 64$.}
\label{tab:number_of_relus_in_each_network}

\end{table}

\paragraph{Networks} We focus on ResNet networks since they are important in the Vision domain. Due to their convolutional property, the number of ReLUs applied for images of shape $32 \times 32$ is smaller than the number of ReLUs applied for images of shape $64\times 64$. Table~\ref{tab:number_of_relus_in_each_network} summarizes the total number of ReLUs in each network with correspondence to the input image sizes. 

\paragraph{Linearization} In this section we focus on Linearization of Neural Networks. We adopt SNL's definition of Linearization. That is, we consider the case in which the network's non-linear operations are limited to ReLUs. Our goal is to take a Neural Network with an abundance of ReLU computations and adhere to a ReLU budget of $B_{\text{target}}$. 

First, we run SNL, a Linearization algorithm, to arrive at some reference budget $B_{\text{ref}} > B_{\text{target}}$, where $B_{\text{ref}}$ is the ReLU budget of the model we start running our algorithm on, and $B_{\text{target}}$ is the ReLU budget we wish our algorithm to arrive at. Then, as a next step, we run our iterative Block Coordinate Descent algorithm on top of it. For comparison completeness, we also run SNL with a target budget of $B_{\text{target}}$ and compare it with our algorithm's result. We further demonstrate that our algorithm is general in the sense that it is agnostic to the type of ReLU replacement functions. Specifically we show that it could be successfully applied on top of AutoRep~\cite{peng2023autorep} and improve its performance.

\paragraph{Hyperparameters} Here we list the hyperparameters for ResNet18 backbone. The full details of running our method on WideResNet22-8 are in the Supplementary Material. Our DRC is 100 ReLUs, ADT of 0.3\% and a maximum of 50 RTs per iteration. For CIFAR-10 and CIFAR-100 we use a finetune routine of 20 epochs per iteration, in contrast to TinyImageNet, for which we require just 5 epochs per iteration. In this case, all finetune routines are performed with SGD~\cite{Kiefer1952StochasticEO} optimizer with an initial learning rate of $10^{-3}$ and Cosine Annealing~\cite{loshchilov2016sgdr} scheduler. We use various $B_\textrm{ref}$ budgets, and the summary of them is specified in the Supplementary. All SNL reference and target models were acquired using the official code with the appropriate hyperparameters.

\subsection{Results}

 \begin{table}[]
\centering
\resizebox{0.92\linewidth}{!}{

\begin{tabular}{|ccc||ccc||ccc}
\hline
\multicolumn{3}{|c||}{\textbf{CIFAR-10}}                                                                                              & \multicolumn{3}{c||}{\textbf{CIFAR-100}}                                                                                             & \multicolumn{3}{c|}{\textbf{TinyImageNet}}                                                                                                                \\ \hline \hline
\multicolumn{1}{|c|}{\begin{tabular}[c]{@{}c@{}}ReLU \\ Budget {[}\#K{]}\end{tabular}} & \multicolumn{1}{c|}{SNL}   & Ours           & \multicolumn{1}{c|}{\begin{tabular}[c]{@{}c@{}}ReLU \\ Budget {[}\#K{]}\end{tabular}} & \multicolumn{1}{c|}{SNL}   & Ours           & \multicolumn{1}{c|}{\begin{tabular}[c]{@{}c@{}}ReLU \\ Budget {[}\#K{]}\end{tabular}} & \multicolumn{1}{c|}{SNL}    & \multicolumn{1}{c|}{Ours}           \\ \hline
\multicolumn{1}{|c|}{6}                                                                & \multicolumn{1}{c|}{84.39} & \textbf{85.65} & \multicolumn{1}{c|}{6}                                                                & \multicolumn{1}{c|}{57.65} & \textbf{61.78} & \multicolumn{1}{c|}{59.1}                                                             & \multicolumn{1}{c|}{54.24} & \multicolumn{1}{c|}{\textbf{56.6}}  \\ \hline
\multicolumn{1}{|c|}{9}                                                                & \multicolumn{1}{c|}{86.97} & \textbf{87.49} & \multicolumn{1}{c|}{9}                                                                & \multicolumn{1}{c|}{62.8}  & \textbf{64.8}  & \multicolumn{1}{c|}{99.6}                                                             & \multicolumn{1}{c|}{58.94}  & \multicolumn{1}{c|}{\textbf{59.77}} \\ \hline
\multicolumn{1}{|c|}{15}                                                               & \multicolumn{1}{c|}{88.43} & \textbf{89.65} & \multicolumn{1}{c|}{15}                                                               & \multicolumn{1}{c|}{67.17} & \textbf{68.85} & \multicolumn{1}{c|}{150}                                                              & \multicolumn{1}{c|}{62.12}  & \multicolumn{1}{c|}{\textbf{62.98}} \\ \hline
\multicolumn{1}{|c|}{20}                                                               & \multicolumn{1}{c|}{90.76}     & \textbf{90.78}     & \multicolumn{1}{c|}{20}                                                               & \multicolumn{1}{c|}{69.38} & \textbf{69.7}  & \multicolumn{1}{c|}{200}                                                              & \multicolumn{1}{c|}{63.39}  & \multicolumn{1}{c|}{\textbf{63.97}} \\ \hline
\multicolumn{1}{|c|}{100}                                                              & \multicolumn{1}{c|}{93.5}  & \textbf{93.78} & \multicolumn{1}{c|}{100}                                                              & \multicolumn{1}{c|}{75.42} & \textbf{76.5}  &                                                                                       &                             &                                     \\ \cline{1-6}
\multicolumn{1}{|c|}{150}                                                              & \multicolumn{1}{c|}{94.26} & \textbf{94.94} & \multicolumn{1}{c|}{150}                                                              & \multicolumn{1}{c|}{76.13} & \textbf{77.27} &                                                                                       &                             &                                     \\ \cline{1-6}
\end{tabular}
}
\caption{\textbf{Test Accuracy[\%] vs. ReLU Budget [\#K] for Wide-ResNet-22-8:} We show that our method outperforms SNL on every budget. }

\label{tab:network_linearization_ours_vs_snl}

\end{table}

\begin{table}[]
\centering
\resizebox{0.92\linewidth}{!}{

\begin{tabular}{ccc||ccc||ccc}
\hline
\multicolumn{3}{|c||}{\textbf{CIFAR-10}}                                                                                            & \multicolumn{3}{c||}{\textbf{CIFAR-100}}                                                                                           & \multicolumn{3}{c|}{\textbf{TinyImageNet}}                                                                                                             \\ \hline
\hline
\multicolumn{1}{|c|}{\begin{tabular}[c]{@{}c@{}}ReLU \\ Budget {[}\#K{]}\end{tabular}} & \multicolumn{1}{c|}{SNL}   & Ours           & \multicolumn{1}{c|}{\begin{tabular}[c]{@{}c@{}}ReLU \\ Budget {[}\#K{]}\end{tabular}} & \multicolumn{1}{c|}{SNL}   & Ours           & \multicolumn{1}{c|}{\begin{tabular}[c]{@{}c@{}}ReLU \\ Budget {[}\#K{]}\end{tabular}} & \multicolumn{1}{c|}{SNL}   & \multicolumn{1}{c|}{Ours}           \\ \hline
\multicolumn{1}{|c|}{50}                                                             & \multicolumn{1}{c|}{92.85} & \textbf{93.21} & \multicolumn{1}{c|}{50}                                                             & \multicolumn{1}{c|}{72.1}  & \textbf{72.6}  & \multicolumn{1}{c|}{200}                                                            & \multicolumn{1}{c|}{58.4}  & \multicolumn{1}{c|}{\textbf{60.5}}  \\ \hline
\multicolumn{1}{|c|}{240}                                                            & \multicolumn{1}{c|}{94.24} & \textbf{95.29} & \multicolumn{1}{c|}{120}                                                            & \multicolumn{1}{c|}{76.35} & \textbf{76.38} & \multicolumn{1}{c|}{250}                                                            & \multicolumn{1}{c|}{60.36} & \multicolumn{1}{c|}{\textbf{61.27}} \\ \hline
\multicolumn{1}{|c|}{300}                                                            & \multicolumn{1}{c|}{95.06} & \textbf{95.37} & \multicolumn{1}{c|}{150}                                                            & \multicolumn{1}{c|}{77.35} & \textbf{77.39} & \multicolumn{1}{c|}{488.8}                                                          & \multicolumn{1}{c|}{64.42} & \multicolumn{1}{c|}{\textbf{65.04}} \\ \hline
                                                                                     &                            & \textbf{}      & \multicolumn{1}{c|}{180}                                                            & \multicolumn{1}{c|}{77.65} & \textbf{78.08} &                                                                                     &                            &                                     \\ \cline{4-6}
\end{tabular}
}
\caption{\textbf{Test Accuracy[\%] vs. ReLU Budget [\#K] for ResNet18:} We show that our method outperforms SNL on every budget. Comparison with previous methods is reported in Figure~\ref{fig:acc_vs_relu_budget_three_datasets} and Figure~\ref{fig:ours_vs_senet}.}

\label{tab:network_linearization_ours_vs_snl_wrn}
\end{table}

\paragraph{Comparing against SNL:}
Figure~\ref{fig:acc_vs_relu_budget_three_datasets} shows the task performance achieved by running our method on ResNet18, compared with previous works for the common benchmark of three classification datasets including CIFAR-10, CIFAR-100 and TinyImageNet. Our method outperforms all other methods in this benchmark.  

Since SNL~\cite{cho2022selective} follows a standard training procedure of ResNets training,  we summarize our performance relative to SNL on ResNet18 backbone in Table~\ref{tab:network_linearization_ours_vs_snl} and on WideResNet22-8 backbone in Table~\ref{tab:network_linearization_ours_vs_snl_wrn}.  We utilize the same architecture as SNL with an alternative set of enabled ReLUs. Therefore, the Private Inference latency figure of our method is the exact same as SNL's in $B_{\text{target}}$. Therefore, we report the network accuracy [\%] vs the ReLU budget [\# ReLUs] metric.

For CIFAR-10, we outperform SNL in some of the very low ReLU budget scenarios, by up to 1.2\%. In the case of CIFAR-100, we achieve better results than SNL in all the tested scenarios, and the gap gets wider as the ReLU budget decreases. We observe an uplift in test accuracy of 4\% in the lowest tested ReLU budget of $6K$ ReLUs. We observe a similar trend to the previous datasets; our method outperforms SNL in all tested budgets. In the extremely low ReLU budget setting, we observe an improvement of up to almost 2.5\%.

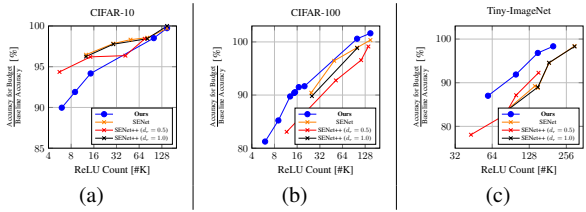
\begin{figure}
    \centering
    \resizebox{0.98\linewidth}{!}{
    \begin{tabular}{c|c|c}
        \resizebox{0.33\linewidth}{!}{\begin{tikzpicture}
    \begin{semilogxaxis}[legend pos=south east,
        legend style={nodes={scale=0.62, transform shape}},
        width=0.7\columnwidth,
        height=5.7cm,
        xlabel= {ReLU Count [\#K]},
        ylabel= {$\frac{\textrm{Accuracy for Budget}}{\textrm{Baseline Accuracy}}$ [\%]},
        xlabel style={at={(0.5, 0.1)}},
        ylabel style={at={(-0.05, 0.5)}},
        title=CIFAR-10,
        xmin=4, xmax=200,
        ymin=85,ymax=100,
        ylabel near ticks,
        xlabel near ticks,
        axis background/.style={fill=blue!0},
        grid=both,
        log basis x = 2,
        /pgf/number format/1000 sep={\,},
        log ticks with fixed point,
        grid style={line width=.1pt, draw=gray!10},
        major grid style={line width=.2pt,draw=gray!50},
        ]

     \addplot[mark=*,mark size=2.5pt,blue, thick]
        plot coordinates {
            (6, 89.97)
            (9, 91.90)
            (14.5, 94.17)
            (100, 98.51)
            (150, 99.73)
        };
    \addlegendentry{\textbf{Ours}}

    \addplot[mark=x,mark size=2.5pt,orange, thick]
        plot coordinates {
            (12.6, 96.47)
            (29, 97.84)
            (49.1, 98.32)
            (82, 98.56)
            (150, 99.70)

        };
    \addlegendentry{SENet}

    \addplot[mark=x,mark size=2.5pt,red, thick]
        plot coordinates {
            (5.6, 94.36)
            (14.4, 96.21)
            (42, 96.36)
            (75.9, 98.43)

        };
    \addlegendentry{SENet++ ($d_r=0.5$)}

    \addplot[mark=x,mark size=2.5pt,black, thick]
        plot coordinates {
            (12.6, 96.25)
            (29, 97.76)
            (82, 98.44)
            (150, 100.0)

        };
    \addlegendentry{SENet++ ($d_r=1.0$)}

    \end{semilogxaxis}
\end{tikzpicture}} & \resizebox{0.33\linewidth}{!}{\begin{tikzpicture}
    \begin{semilogxaxis}[legend pos=south east,
        legend style={nodes={scale=0.62, transform shape}},
        width=0.7\columnwidth,
        height=5.7cm,
        xlabel={ReLU Count [\#K]},
        ylabel={$\frac{\textrm{Accuracy for Budget}}{\textrm{Baseline Accuracy}}$ [\%]},
        xlabel style={at={(0.5, 0.1)}},
        ylabel style={at={(-0.05, 0.5)}},
        title=CIFAR-100,
        xmin=4, xmax=200,
        ymin=80,ymax=103,
        ylabel near ticks,
        xlabel near ticks,
        axis background/.style={fill=blue!0},
        grid=both,
        log basis x = 2,
        /pgf/number format/1000 sep={\,},
        log ticks with fixed point,
        grid style={line width=.1pt, draw=gray!10},
        major grid style={line width=.2pt,draw=gray!50},
        ]

     \addplot[mark=*,mark size=2.5pt,blue, thick]
        plot coordinates {
            (6, 81.25 )
            (9, 85.22 )
            (12.9, 89.74 )
            (14.6, 90.40 )
            (15, 90.54 )
            (16.8, 91.49 )
            (20, 91.66 )
            (100, 100.60 )
            (150, 101.62 )
        };
    \addlegendentry{\textbf{Ours}}

    \addplot[mark=x,mark size=2.5pt,orange, thick]
        plot coordinates {
            (24.6, 90.44 )
            (49.6, 96.45 )
            (100, 98.93 )
            (150, 100.35 )

        };
    \addlegendentry{SENet}

    \addplot[mark=x,mark size=2.5pt,red, thick]
        plot coordinates {
            (11.6, 83.08)
            (52.2, 92.76)
            (113, 96.57)
            (141, 99.14)

        };
    \addlegendentry{SENet++ ($d_r=0.5$)}

    \addplot[mark=x,mark size=2.5pt,black, thick]
        plot coordinates {
            (25.1, 89.81)
            (100, 98.83)

        };
    \addlegendentry{SENet++ ($d_r=1.0$)}

    \end{semilogxaxis}
\end{tikzpicture}} & \resizebox{0.33\linewidth}{!}{\begin{tikzpicture}
    \begin{semilogxaxis}[legend pos=south east,
        legend style={nodes={scale=0.62, transform shape}},
        width=0.7\columnwidth,
        height=5.7cm,
        xlabel={ReLU Count [\#K]},
        ylabel={$\frac{\textrm{Accuracy for Budget}}{\textrm{Baseline Accuracy}}$ [\%]},
        xlabel style={at={(0.5, 0.1)}},
        ylabel style={at={(-0.05, 0.5)}},
        title=Tiny-ImageNet,
        xmin=32, xmax=350,
        ymin=75,ymax=103,
        ylabel near ticks,
        xlabel near ticks,
        axis background/.style={fill=blue!0},
        grid=both,
        log basis x = 2,
        /pgf/number format/1000 sep={\,},
        log ticks with fixed point,
        grid style={line width=.1pt, draw=gray!10},
        major grid style={line width=.2pt,draw=gray!50},
        ]

     \addplot[mark=*,mark size=2.5pt,blue, thick]
        plot coordinates {
            (59.1, 87.01)
            (99.6, 91.88)
            (150, 96.82)
            (200, 98.34)
        };
    \addlegendentry{\textbf{Ours}}

    \addplot[mark=x,mark size=2.5pt,orange, thick]
        plot coordinates {
            (80, 83.23)
            (142, 89.23)
            (150, 88.8)
            (185, 94.52)
            (298, 98.41)
        };
    \addlegendentry{SENet}

    \addplot[mark=x,mark size=2.5pt,red, thick]
        plot coordinates {
            (43, 78.09)
            (81, 82.63)
            (100, 87.14)
            (152, 92.28)

        };
    \addlegendentry{SENet++ ($d_r=0.5$)}

    \addplot[mark=x,mark size=2.5pt,black, thick]
        plot coordinates {
            (80, 83.17)
            (150, 88.98)
            (185, 94.58)
            (298, 98.3)
        };
    \addlegendentry{SENet++ ($d_r=1.0$)}

    \end{semilogxaxis}
\end{tikzpicture}} \\
        (a) & (b) & (c)
    \end{tabular}
    }
    \caption{\textbf{Ours vs SENet on a ResNet18 backbone:} We show that our method achieves the Pareto frontier on network accuracy with respect to ReLU budget on CIFAR-100 and TinyImageNet, while staying competitive on CIFAR-10 against SENet and SENet++. We test using a metric which is agnostic to the baseline classifier accuracy. Specifically, we measure the ratio between the performance reached by executing each method and divide it by the accuracy of the baseline. Namely: $\frac{\textrm{Accuracy for Budget}}{\textrm{Baseline Accuracy}}$ }
    \label{fig:ours_vs_senet}
\end{figure}

\paragraph{Comparing against SENet:}
Since SENet~\cite{kundu2023senet} uses baseline models which are typically at higher accuracies than SNL and AutoRep, we report comparisons to SENet in a metric which is agnostic to the baseline model accuracy. Figure~\ref{fig:ours_vs_senet} shows that we surpass SENet for all cases in the CIFAR-100 and TinyImageNet datasets and that our method is competitive with SENet in the case of CIFAR-10. Graphs comparing our method and SENet on the WideResNet22-8 backbone can be found in the Supplemental.

\paragraph{Comparing against AutoReP:}
Our algorithm can be applied to generate better ReLU-sparse models on top of additional methods. We demonstrate that our optimization scheme is general and can work also with AutoRep~\cite{peng2023autorep}. The latter suggests replacing ReLUs with second order polynomial approximations of ReLUs. We show that starting from a baseline AutoRep network, we can arrive at better networks at lower ReLU budgets.

We test the CIFAR-100 setting for both ResNet18 and WideResNet22-8 backbones. We start from some AutoRep model as our baseline. Then, we invoke our algorithm on that model to arrive at very low budgets. We refer the reader for the Supplementary Material for the specifications of hyperparameters. 

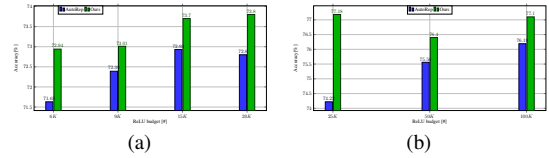
\begin{figure}
    \centering
    \resizebox{0.9\linewidth}{!}{
    \begin{tabular}{cc}
       \resizebox{0.5\linewidth}{!}{\begin{tikzpicture}
    \begin{axis}[
            ybar,
            bar width=.5cm,
            width=\textwidth,
            height=.5\textwidth,
            legend style={at={(0.2,1)},
                anchor=north,legend columns=-1},
            symbolic x coords={$6K$,$9K$,$15K$,$20K$},
            xtick=data,
            nodes near coords,
            nodes near coords align={vertical},
            xlabel={ReLU budget [\#]},
            ylabel={Accuracy[\%]},
            grid=both,
        ]
        \addplot[blue!20!black,fill=blue!80!white] table[x=interval,y=autorep]{\mydata};
        \addplot[green!20!black,fill=green!70!black] table[x=interval,y=ours]{\mydata};
        \legend{AutoRep, Ours}
        
    \end{axis}
\end{tikzpicture}}  & \resizebox{0.5\linewidth}{!}{\begin{tikzpicture}
    \begin{axis}[
            ybar,
            bar width=.5cm,
            width=\textwidth,
            height=.5\textwidth,
            legend style={at={(0.5,1)},
                anchor=north,legend columns=-1},
            symbolic x coords={$25K$,$50K$,$100K$,},
            xtick=data,
            nodes near coords,
            nodes near coords align={vertical},
            xlabel={ReLU budget [\#]},
            ylabel={Accuracy[\%]},
            grid=both,
        ]
        \addplot[blue!20!black,fill=blue!80!white] table[x=interval,y=autorep]{\mydataWRN};
        \addplot[green!20!black,fill=green!70!black] table[x=interval,y=ours]{\mydataWRN};
        \legend{AutoRep, Ours}
        
    \end{axis}
\end{tikzpicture}}
\\
         (a) & (b)
    \end{tabular}
    
}
    \caption{\textbf{Our method on top of AutoRep for CIFAR-100: } (a) for a ResNet18 backbone, (b) for a WideResNet22-8 backbone. Our method can be plugged on top of any Selective method. Here we demonstrate the performance gain using our method on top of AutoRep. In the case of ResNet18, our method achieves 72.9\% accuracy with 6K ReLUs, compared with AutoRep which achieves this performance with more than $\times 2$ the number of ReLUs (15K). The same trend occurs for WideResNet22-8 as well, where 76\% accuracy is achieved in our case with \emph{half} of the ReLU budget that this accuracy is achieved in the case of AutoRep.}
    \label{fig:autorep_resnet18_cifar100_comparison}

\end{figure}

Figure~\ref{fig:autorep_resnet18_cifar100_comparison} summarizes these results. We show that for all cases, using our method improves AutoRep's performance. We utilize the same architecture as AutoRep with an alternative set of enabled ReLUs. Therefore, the Private Inference latency figure of our method is the exact same as AutoRep's in $B_{\text{target}}$.


\pgfplotstableread[row sep=\\,col sep=&]{
    interval & autorep & ours \\
    $6K$     & 71.63  & 72.94  \\
    $9K$     & 72.39 & 73.01  \\
    $15K$    & 72.93 & 73.7 \\
    $20K$   & 72.8 & 73.8 \\
    }\mydata

\pgfplotstableread[row sep=\\,col sep=&]{
    interval & autorep & ours \\
    $25K$     & 74.22  & 77.18  \\
    $50K$     & 75.56 & 76.4  \\
    $100K$    & 76.19 & 77.1 \\
    }\mydataWRN

\section{Discussion}

\paragraph{Ours vs. SNL:}
We formally state the problem of optimizing a neural network while adhering to a specific ReLU budget in Equation~\ref{eq:general_case_subject_to}. Selective approaches such as Network Linearization (SNL)~\cite{cho2022selective} suggest to replace the constraint $||m ||_0 \leq B$ with a regularization term which is governed by a Lagrange multiplier. To achieve that, they perform two relaxations with respect to the optimization problem in Equation~\ref{eq:general_case_subject_to}: first, they replace the $L_0$ constraint with an $L_1$ regularization term. Next, they introduce a hyperparameter $\lambda$, also known as Lagrange Multiplier or termed Lasso coefficient. This is formally stated as:
\begin{equation}
    \begin{aligned}
    \min_{\theta, m} \quad & \mathcal{L}_{\text{cross-entropy}}(f_{\theta, m }(x), y) + \lambda \cdot ||m ||_1\\
    \end{aligned}
    \label{eq:snls_optimization}
\end{equation}

Selective methods make two relaxations: first they replace $||\cdot ||_0$ with $||\cdot ||_1$, second they select the extent to which CE-loss is preferred over the regularization term through choosing a specific value for $\lambda$. Our method on the other hand, works directly on the binary mask $m$, therefore we work on the space measured by the $L_0$ norm without the $L_1$ norm relaxation. Our straightforward, relaxation-free approach is demonstrated to achieve better network performance relative to existing Selective methods.

\paragraph{Ours vs. AutoReP:} AutoRep's hysteresis mechanism demonstrates the effort one needs to take to stabilize a Selective approach solution. The reason such measure should take place is that Selective approaches use SGD. The problem is that SGD does not produce sparse solution by-design in the formulation of Equation~\ref{eq:snls_optimization}. 

We, on the other hand, use Block Coordinate Descent which offers a sparse solution by-design. Moreover, every state of our optimization scheme yields a sparse solution. It has been shown~\cite{shalev2009stochastic} that the relationship between the task performance of the optimal solution and task performance of the solution suggested by a Block Coordinate Descent method is inverse proportional to the number of iterations. Given a mask of dimension $d$, an optimal map $m^{*}$, if we use Block Coordinate Descent, for a ReLUs map $m_T$ obtained after $T$ steps, we have that:

\begin{equation}
    \label{eq:sss_prob}
    E[P(m_T)] - P(m^{*}) \leq \frac{d \cdot \Psi(0)}{T + 1}
\end{equation}

where $\Psi(m)$ is:
\begin{equation}
    \label{eq:big_psi}
   \Psi(m) = \frac{\beta}{2} \cdot ||m^{*} - m||_2^2 +P(m)
\end{equation}
and $P(m)$ is:
\begin{equation}\label{eq:p_of_m}
    P(m) = \mathcal{L}_{\text{cross-entropy}}(f_{\theta, m}(x), y) + \lambda \cdot ||m||_1
\end{equation}
Furthermore, with probability larger than 0.5, we have that:
\begin{equation}
    \label{eq:sss}
    P(m_T) - P(m^{*}) \leq \frac{2 \cdot d \cdot \Psi(0)}{T + 1}
\end{equation}

The term $P(m_T) - P(m^{*})$ denotes the gap between the optimal algorithm's task performance and the performance of a Block Coordinate Descent method. Therefore, Equations~\ref{eq:sss} and ~\ref{eq:sss_prob} are provable guarantees on the task performance of a Block Coordinate Descent method with respect to the number of iteration steps $T$.

This provable guarantee correlates well with our ablation study that shows the relationship between task performance and DRC in Figure~\ref{fig:hyperparams_comparisons}(a). DRC is inverse proportional to the number of steps $T$: $T = \lceil\frac{B_\textrm{ref} - B_\textrm{target}}{DRC}\rceil$. We observe a decrease in network performance for increasing values of DRC, conversely there is an increase in network performance for increasing values of the iteration steps as reflected in the analysis by~\cite{shalev2009stochastic}.

\section{Ablation Study}
First, we cover the effect of choosing different hyperparameters for our network. Then, we turn to an intuitive justification for using Block Coordinate Descent through an analysis of the Dynamics of ReLU masks. Finally, we show the ReLU distribution across layers in networks that are produced by our approach.

\subsection{Hyperparameters Study} \label{subsec:hyperparameter_study}

We study the effect of our algorithm's hyperparameters on its performance; namely, we test different options for the reduce step, the reduce threshold and the number of finetune epochs. 
Our tests are based on the ResNet18/CIFAR-100 Linearization setting with a reference budget of $30K$, and $15K$ ReLUs as the target budget. The baseline configuration for the hyperparameters used here is the same that was described in the experiments section.

\begin{figure*}[]
    \centering
    \resizebox{0.95\linewidth}{!}{
    \begin{tabular}{c c c} 
    \resizebox{0.45\linewidth}{!} {    \begin{tikzpicture}
    \begin{axis}[legend pos=south east,
        legend style={nodes={scale=0.62, transform shape}},
        width=0.7\columnwidth,
        height=5.7cm,
        xlabel={Delta ReLU Count [\#]},
        ylabel={Accuracy [\%]},
        xlabel style={at={(0.5, 0.1)}},
        ylabel style={at={(-0.05, 0.5)}},
        title={Accuracy[\%] vs. DRC[\#]},
        xmin=50, xmax=550,
        ymin=60,ymax=75,
        ylabel near ticks,
        xlabel near ticks,
        axis background/.style={fill=blue!0},
        grid=both,
        /pgf/number format/1000 sep={\,},
        grid style={line width=.1pt, draw=gray!10},
        major grid style={line width=.2pt,draw=gray!50},
        ]
     \addplot[mark=*,mark size=2.5pt,blue, thick]
        plot coordinates {
            (100, 68.89)
            (200, 66.93)
            (300, 64.89)
            (500, 62.68)
        };
    	
    \addplot[mark=o,mark size=5pt,orange, thick]
        plot coordinates {
            (100, 68.89)
        };

    \end{axis}
\end{tikzpicture}}
          &
         \resizebox{0.55\linewidth}{!}{\begin{tikzpicture}
    \begin{axis}[legend pos=south east,
        legend style={nodes={scale=0.62, transform shape}},
        width=0.7\columnwidth,
        height=5.7cm,
        xlabel={Accuracy Degradation Tolerance (ADT) [\%]},
        ylabel={Accuracy [\%]},
        xlabel style={at={(0.5, 0.1)}},
        ylabel style={at={(-0.05, 0.5)}},
        title={\begin{tabular}{c}
           {Accuracy[\%] vs.}\\
           {Accuracy Degradation Tolerance (ADT)[\%]}
        \end{tabular}},
        xmin=0, xmax=0.6,
        ymin=65,ymax=72,
        ylabel near ticks,
        xlabel near ticks,
        axis background/.style={fill=blue!0},
        grid=both,
        /pgf/number format/1000 sep={\,},
        grid style={line width=.1pt, draw=gray!10},
        major grid style={line width=.2pt,draw=gray!50},
        ]
     \addplot[mark=*,mark size=2.5pt,blue, thick]
        plot coordinates {
            (0.1, 68.61)
            (0.2, 68.67)
            (0.3, 68.89)
            (0.5, 68.66)
        };
    	
    \addplot[mark=o,mark size=5pt,orange, thick]
        plot coordinates {
            (0.3, 68.89)
        };

    \end{axis}
\end{tikzpicture}} &
        \resizebox{0.45\linewidth}{!}{\begin{tikzpicture}
    \begin{axis}[legend pos=south east,
        legend style={nodes={scale=0.62, transform shape}},
        width=0.7\columnwidth,
        height=5.7cm,
        xlabel={Finetune Epochs [\#]},
        ylabel={Accuracy [\%]},
        xlabel style={at={(0.5, 0.1)}},
        ylabel style={at={(-0.05, 0.5)}},
        title={\begin{tabular}{c}
           {Accuracy[\%] vs.}\\
           {Finetune Epochs [\#]}
        \end{tabular}},
        xmin=1, xmax=35,
        ymin=60,ymax=75,
        ylabel near ticks,
        xlabel near ticks,
        axis background/.style={fill=blue!0},
        grid=both,
        /pgf/number format/1000 sep={\,},
        grid style={line width=.1pt, draw=gray!10},
        major grid style={line width=.2pt,draw=gray!50},
        ]
     \addplot[mark=*,mark size=2.5pt,blue, thick]
        plot coordinates {
            (5, 63.63)
            (10, 66.23)
            (20, 68.89)
            (30, 68.29)
        };
    	
    \addplot[mark=o,mark size=5pt,orange, thick]
        plot coordinates {
            (20, 68.89)
        };

    \end{axis}
\end{tikzpicture}}
         \\
         (a) & (b) & (c) \\
    \end{tabular}
    }
    \caption{\textbf{Hyperparameters ablation study:} We compare the effect of various hyperparameters on the performance of the algorithm on the ResNet18/CIFAR-100 setting. Chosen parameters are circled. (a) DRC is inverse proportional to the number of iteration steps $T = \lceil\frac{B_\textrm{ref} - B_\textrm{target}}{DRC}\rceil$. Accuracy decreases as the DRC increases. Conversely, the accuracy increases as the number of iteration steps $T$ increases, as anticipated in Equations~\ref{eq:sss_prob} and~\ref{eq:sss}. In (b) and (c) we evaluate the effect of the number of finetune epochs and the Accuracy Degradation Tolerance (ADT) on the test accuracy. }
    \label{fig:hyperparams_comparisons}
\end{figure*}
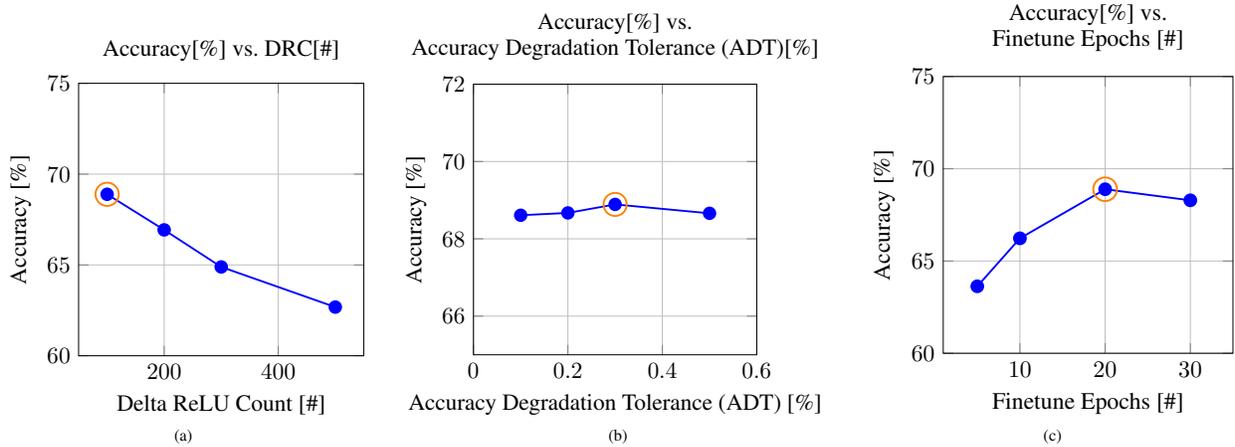

Figure~\ref{fig:hyperparams_comparisons}(a) shows that the reduce step has a crucial effect on performance, especially when working with low budgets. It should be noted that when working with higher ReLU budgets, the reduce step may be tuned to be higher to speed up the algorithm.

Figure~\ref{fig:hyperparams_comparisons}(b) shows  that the number of finetune epochs has a significant bearing on the resulting model's performance. It should be noted that we also observe diminishing returns for a high enough number of finetune epochs and that is expected, as at some point, the learning rate decays too much due to the behavior of the scheduler, and thus not improving the model's performance.

Figure~\ref{fig:hyperparams_comparisons}(c) shows that the Accuracy Degradation Tolerance (ADT) does not affect the model's performance in a significant way in this setting. It is worth pointing out that this hyperparameter allows for optimizing run time such that there is no need to perform additional ReLU selections and reduction attempts at the same step. However, great care must be taken to not choose a too high value so that model is able to keep up with the reduction after finetuning it.


\subsection{ReLU mask Dynamics for Decreasing Budgets}

Denote $m_1^{*}$ as the optimal solution mask for budget $B_1$ and $m_2^{*}$ for $B_2$. Assume that $B_2 > B_1$. \textbf{What is the relationship between $m_1^{*}$ and $m_2^{*}$?} On one extreme case, the masks are totally disjoint. If there are common ReLUs between the two masks, how many are there?

Our algorithm assumes that there is a complete overlap between the two. But is there evidence to this assumption in approaches which do not assume this by design? We take a single SNL optimization path. We record the snapshot of masks along the optimization process from the full network capacity until the target budget. 

For a target budget of $30K$ on CIFAR-100 and ResNet18, SNL runs for about $300$ epochs. This gives about ${300 \choose 2} \simeq 45K$ pairs of $(m_1, m_2)$ corresponding to budgets $(B_1, B_2)$ such that $B_2 > B_1$. For each pair, we evaluate an Intersection over Union (IoU) score: $\frac{||m_1 \odot m_2||_0}{ || m_1||_0} \in [0, 1]$ where $\odot$ denotes element-wise product. This score quantifies the overlap between the two masks.

\begin{figure}
    \centering
    \begin{tabular}{c|c}
   IoU score vs. epoch &  IoU score of ($B_1$, $B_2$)\\
           \includegraphics[width=0.49\linewidth]{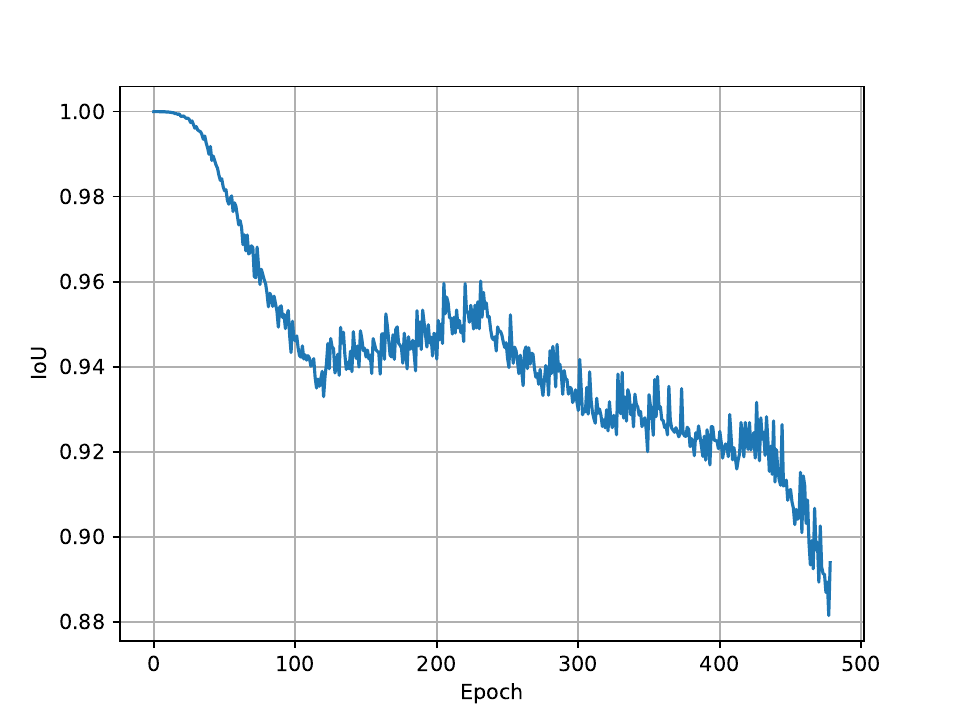}
           &
           \includegraphics[width=0.45\linewidth]{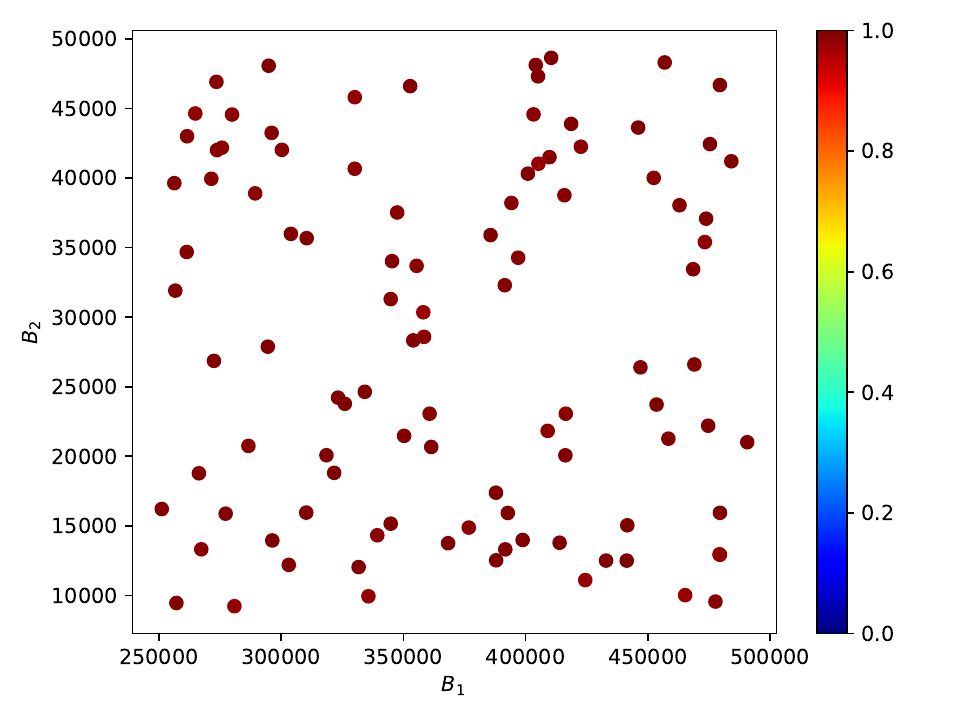} \\
         (a) & (b) \\
    \end{tabular}
    \caption{\textbf{ReLU masks IoU: } (a) Shows the IoU of consecutive binarized ReLU masks over epochs during SNL. (b) Each point represents two ReLU masks from two ReLU budget checkpoints $(B_1, B_2)$ along a Selective training procedure, we color code the IoU score of the two ReLU masks. All points are red because the IoU between $B_1$ and $B_2$ is consistently higher than 0.85.}
    \label{fig:scatter_plot_masks_iou}
\end{figure}

Figure~\ref{fig:scatter_plot_masks_iou} shows two plots of the IoU scores. Figure~\ref{fig:scatter_plot_masks_iou}(a) shows the IoU scores for every two consecutive ReLU budgets during the optimization process. We can observe that the IoU scores are well above $0.85$. Figure~\ref{fig:scatter_plot_masks_iou}(b) shows a random sample of budgets and their corresponding masks, through the entire optimization process. Observe that all points are colored red indicating that the IoU score is above $0.85$. 
We conclude that the number of shared ReLUs between the masks is high and therefore the majority of ReLUs changed from one mask to another is with ReLU elimination as opposed to ReLU addition.

This motivates us to assume that a golden set of ReLUs exists. We conjecture that an algorithm which eliminates ReLUs consecutively by design can reach a solution close to optimal. Therefore we propose our optimization strategy which by design eliminates ReLUs without revisiting them in subsequent iteration steps.

\subsection{ReLU distribution of Our network} 
\begin{figure}
\resizebox{0.98\linewidth}{!}{
\begin{tabular}{c|c}
    {

\pgfplotstableread[row sep=\\,col sep=&]{
    interval & full & ours \\
    0 & 65536 & 0 \\
    1 & 65536 & 73 \\
    2 & 65536 & 721 \\
    3 & 65536 & 13 \\
    4 & 65536 & 3063 \\
    5 & 32768 & 790 \\
    6 & 32768 & 589 \\
    7 & 32768 & 477 \\
    8 & 32768 & 111 \\
    9 & 16384 & 1635 \\
    10 & 16384 & 246 \\
    11 & 16384 & 546 \\
    12 & 16384 & 16 \\
    13 & 8192 & 1894 \\
    14 & 8192 & 2908 \\
    15 & 8192 & 1620 \\
    16 & 8192 & 294 \\
    }\mydata

\begin{tikzpicture}
    \begin{axis}[
            ybar,
            bar width=.15cm,
            width=\textwidth,
            height=.5\textwidth,
            legend style={at={(0.7,1)},
                anchor=north ,legend columns=-1},
            symbolic x coords={0, 1, 2, 3, 4, 5, 6, 7, 8, 9, 10, 11, 12, 13, 14, 15, 16},
            xtick=data,
            xlabel={Layer Index},
            ylabel={ReLU Count [\#]},
            ymode=log,
        ]
        \addplot[blue!20!black,fill=blue!80!white,opacity=0.5] table[x=interval,y=full]{\mydata};
        \addplot[green!20!black,fill=green!70!black,opacity=0.5] table[x=interval,y=ours]{\mydata};
        \legend{Full Model, Ours}
    
    \end{axis}
\end{tikzpicture}} & 
    {

\pgfplotstableread[row sep=\\,col sep=&]{
    interval & snlRef & snlTarget & ours \\
1 & 162 & 119 & 73 \\ 
2 & 1465 & 1011 & 721 \\ 
3 & 33 & 25 & 13 \\ 
4 & 6187 & 4902 & 3063 \\ 
5 & 1611 & 822 & 790 \\ 
6 & 1122 & 563 & 589 \\ 
7 & 974 & 458 & 477 \\ 
8 & 223 & 185 & 111 \\ 
9 & 3321 & 2197 & 1635 \\ 
10 & 509 & 124 & 246 \\ 
11 & 1079 & 646 & 546 \\ 
12 & 33 & 18 & 16 \\ 
13 & 3736 & 1128 & 1894 \\ 
14 & 5749 & 1249 & 2908 \\ 
15 & 3111 & 922 & 1620 \\ 
16 & 581 & 252 & 294 \\ 
    }\mydata

\begin{tikzpicture}
    \begin{axis}[
            ybar,
            bar width=.07cm,
            width=\textwidth,
            height=.5\textwidth,
            legend style={at={(0.7,1)},
                anchor=north ,legend columns=-1},
            symbolic x coords={1, 2, 3, 4, 5, 6, 7, 8, 9, 10, 11, 12, 13, 14, 15, 16},
            xtick=data,
            xlabel={Layer Index},
            ylabel={ReLU Count [\#]},
            ymode=log,
        ]
        \addplot[red!20!black,fill=red!80!white,opacity=0.5] table[x=interval,y=snlRef]{\mydata};
        \addplot[blue!20!black,fill=blue!80!white,opacity=0.5] table[x=interval,y=snlTarget]{\mydata};
        \addplot[green!20!black,fill=green!70!black,opacity=0.5] table[x=interval,y=ours]{\mydata};
        \legend{SNL@$B_{\textrm{ref}}$, SNL@$B_{\textrm{target}}$, Ours}
    
    \end{axis}
\end{tikzpicture}}
    \\
    (a)  & (b) \\
\end{tabular}
}
    \caption{\textbf{ReLU distribution across layers: } (a) shows the original ReLU distribution in blue and our ReLU distribution result in green. (b) shows the ReLU distribution of SNL in the reference budget $B_{\textrm{ref}}=30K$ in red, SNL at $B_{\textrm{target}}=15K$ in blue and our method in green. Note that layer \#0 is ReLU free in all cases and is therefore excluded from the graph.}
    \label{fig:relu_distribution}
\end{figure}
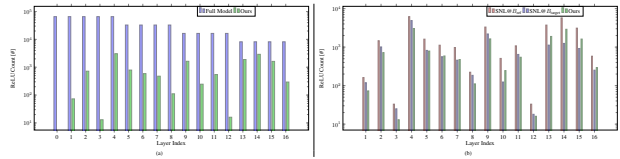

In this section we show the distribution of ReLUs that our method decided to keep. We refer to ResNet18/CIFAR-100 setting with a target ReLU budget of $15K$ ReLUs.
Figure~\ref{fig:relu_distribution}(a) shows the distribution of ReLUs across the different layers with respect to the original ResNet18 network. Figure~\ref{fig:relu_distribution}(b) shows the distribution of ReLUs across the different layers for 3 methods: our method, SNL at the reference budget and SNL at the target budget.

\section{Conclusion}
We addressed the problem of Private Inference and, in particular, focused on reducing the number of ReLUs that form a bottleneck in PI applications. This is a discrete problem where the goal is to remove as many ReLUs as possible, with as little impact on accuracy as possible. Current state-of-the-art methods rely on relaxation techniques to optimize this problem, but we, instead, use Block Coordinate Descent for the optimization. Our key insight lies in that we treat the ReLU reduction problem as a \emph{sparse} optimization problem and we treat it as such.

Our method outperforms existing methods in terms of network accuracy, and can be applied on top of existing methods. Specifically, we improve on CIFAR-100 and ResNet18 by $\sim4\%$ on a $6K$ ReLU budget compared with SNL. Moreover, for extremely low budgets, we achieve AutoRep's test accuracy with half of the ReLU budget AutoRep utilizes.

\bibliography{aaai25}

@article{tibshirani1996regression,
  title={Regression shrinkage and selection via the lasso},
  author={Tibshirani, Robert},
  journal={Journal of the Royal Statistical Society Series B: Statistical Methodology},
  volume={58},
  number={1},
  pages={267--288},
  year={1996},
  publisher={Oxford University Press}
}

@inproceedings{srinivasan2019delphi,
  title={DELPHI: A cryptographic inference service for neural networks},
  author={Srinivasan, Wenting Zheng and Akshayaram, PMRL and Ada, Popa Raluca},
  booktitle={Proc. 29th USENIX Secur. Symp},
  pages={2505--2522},
  year={2019}
}

@inproceedings{cho2022selective,
  title={Selective network linearization for efficient private inference},
  author={Cho, Minsu and Joshi, Ameya and Reagen, Brandon and Garg, Siddharth and Hegde, Chinmay},
  booktitle={International Conference on Machine Learning},
  pages={3947--3961},
  year={2022},
  organization={PMLR}
}

@inproceedings{peng2023autorep,
  title={Autorep: Automatic relu replacement for fast private network inference},
  author={Peng, Hongwu and Huang, Shaoyi and Zhou, Tong and Luo, Yukui and Wang, Chenghong and Wang, Zigeng and Zhao, Jiahui and Xie, Xi and Li, Ang and Geng, Tony and others},
  booktitle={Proceedings of the IEEE/CVF International Conference on Computer Vision},
  pages={5178--5188},
  year={2023}
}

@inproceedings{shalev2009stochastic,
  title={Stochastic methods for l 1 regularized loss minimization},
  author={Shalev-Shwartz, Shai and Tewari, Ambuj},
  booktitle={Proceedings of the 26th Annual International Conference on Machine Learning},
  pages={929--936},
  year={2009}
}

@article{ghodsi2020cryptonas,
  title={Cryptonas: Private inference on a relu budget},
  author={Ghodsi, Zahra and Veldanda, Akshaj Kumar and Reagen, Brandon and Garg, Siddharth},
  journal={Advances in Neural Information Processing Systems},
  volume={33},
  pages={16961--16971},
  year={2020}
}

@article{cho2022sphynx,
  title={Sphynx: A Deep Neural Network Design for Private Inference},
  author={Cho, Minsu and Ghodsi, Zahra and Reagen, Brandon and Garg, Siddharth and Hegde, Chinmay},
  journal={IEEE Security \& Privacy},
  volume={20},
  number={5},
  pages={22--34},
  year={2022},
  publisher={IEEE}
}

@inproceedings{jha2021deepreduce,
  title={Deepreduce: Relu reduction for fast private inference},
  author={Jha, Nandan Kumar and Ghodsi, Zahra and Garg, Siddharth and Reagen, Brandon},
  booktitle={International Conference on Machine Learning},
  pages={4839--4849},
  year={2021},
  organization={PMLR}
}

@article{juvekar2018gazelle,
  author       = {Chiraag Juvekar and
                  Vinod Vaikuntanathan and
                  Anantha P. Chandrakasan},
  title        = {Gazelle: {A} Low Latency Framework for Secure Neural Network Inference},
  journal      = {CoRR},
  volume       = {abs/1801.05507},
  year         = {2018},
}

@misc{wagh2018securenn,
      author = {Sameer Wagh and Divya Gupta and Nishanth Chandran},
      title = {SecureNN: Efficient and Private Neural Network Training},
      howpublished = {Cryptology ePrint Archive, Paper 2018/442},
      year = {2018},
}

@article{wagh2020falcon,
  author       = {Sameer Wagh and
                  Shruti Tople and
                  Fabrice Benhamouda and
                  Eyal Kushilevitz and
                  Prateek Mittal and
                  Tal Rabin},
  title        = {{FALCON:} Honest-Majority Maliciously Secure Framework for Private
                  Deep Learning},
  journal      = {CoRR},
  volume       = {abs/2004.02229},
  year         = {2020},
}

@article{shamir1979ss,
  author       = {Adi Shamir},
  title        = {How to Share a Secret},
  journal      = {Commun. {ACM}},
  volume       = {22},
  number       = {11},
  pages        = {612--613},
  year         = {1979},
  doi          = {10.1145/359168.359176},
}

@inproceedings{yao1982gc,
  author       = {Andrew Chi{-}Chih Yao},
  title        = {Protocols for Secure Computations (Extended Abstract)},
  booktitle    = {23rd Annual Symposium on Foundations of Computer Science, Chicago,
                  Illinois, USA, 3-5 November 1982},
  pages        = {160--164},
  publisher    = {{IEEE} Computer Society},
  year         = {1982},
  doi          = {10.1109/SFCS.1982.38},
}

@inproceedings{lou2020safenet,
  title={Safenet: A secure, accurate and fast neural network inference},
  author={Lou, Qian and Shen, Yilin and Jin, Hongxia and Jiang, Lei},
  booktitle={International Conference on Learning Representations},
  year={2020}
}

@article{mohassel2017secureml,
  author       = {Payman Mohassel and
                  Yupeng Zhang},
  title        = {SecureML: {A} System for Scalable Privacy-Preserving Machine Learning},
  journal      = {{IACR} Cryptol. ePrint Arch.},
  pages        = {396},
  year         = {2017},
}

@inproceedings{bachrach2016cryptonets,
  author       = {Ran Gilad{-}Bachrach and
                  Nathan Dowlin and
                  Kim Laine and
                  Kristin E. Lauter and
                  Michael Naehrig and
                  John Wernsing},
  editor       = {Maria{-}Florina Balcan and
                  Kilian Q. Weinberger},
  title        = {CryptoNets: Applying Neural Networks to Encrypted Data with High Throughput
                  and Accuracy},
  booktitle    = {Proceedings of the 33nd International Conference on Machine Learning,
                  {ICML} 2016, New York City, NY, USA, June 19-24, 2016},
  series       = {{JMLR} Workshop and Conference Proceedings},
  volume       = {48},
  pages        = {201--210},
  publisher    = {JMLR.org},
  year         = {2016},
}

@inproceedings{he2016deep,
  title={Deep residual learning for image recognition},
  author={He, Kaiming and Zhang, Xiangyu and Ren, Shaoqing and Sun, Jian},
  booktitle={Proceedings of the IEEE conference on computer vision and pattern recognition},
  pages={770--778},
  year={2016}
}

@article{zagoruyko2016wide,
  title={Wide residual networks},
  author={Zagoruyko, Sergey and Komodakis, Nikos},
  journal={arXiv preprint arXiv:1605.07146},
  year={2016}
}

@inproceedings{kundu2023senet,
  author       = {Souvik Kundu and
                  Shunlin Lu and
                  Yuke Zhang and
                  Jacqueline Tiffany Liu and
                  Peter A. Beerel},
  title        = {Learning to Linearize Deep Neural Networks for Secure and Efficient
                  Private Inference},
  booktitle    = {The Eleventh International Conference on Learning Representations,
                  {ICLR} 2023, Kigali, Rwanda, May 1-5, 2023},
  year         = {2023},
}

@article{krizhevsky2009learning,
  title={Learning multiple layers of features from tiny images},
  author={Krizhevsky, Alex and Hinton, Geoffrey and others},
  year={2009},
  publisher={Toronto, ON, Canada}
}

@article{le2015tiny,
  title={Tiny imagenet visual recognition challenge},
  author={Le, Ya and Yang, Xuan},
  journal={CS 231N},
  volume={7},
  number={7},
  pages={3},
  year={2015}
}

@article{Kiefer1952StochasticEO,
  title={Stochastic Estimation of the Maximum of a Regression Function},
  author={J. Kiefer and Jacob Wolfowitz},
  journal={Annals of Mathematical Statistics},
  year={1952},
  volume={23},
  pages={462-466},
}

@article{loshchilov2016sgdr,
  author       = {Ilya Loshchilov and
                  Frank Hutter},
  title        = {{SGDR:} Stochastic Gradient Descent with Restarts},
  journal      = {CoRR},
  volume       = {abs/1608.03983},
  year         = {2016},
}

@inproceedings{kingma2015adam,
  author       = {Diederik P. Kingma and
                  Jimmy Ba},
  title        = {Adam: {A} Method for Stochastic Optimization},
  booktitle    = {3rd International Conference on Learning Representations, {ICLR} 2015,
                  San Diego, CA, USA, May 7-9, 2015, Conference Track Proceedings},
  year         = {2015},
  url          = {http://arxiv.org/abs/1412.6980},
}

@article{hinton2015distilling,
  title={Distilling the knowledge in a neural network},
  author={Hinton, Geoffrey and Vinyals, Oriol and Dean, Jeff},
  journal={arXiv preprint arXiv:1503.02531},
  year={2015}
}

@book{gentry2009fully,
  title={A fully homomorphic encryption scheme},
  author={Gentry, Craig},
  year={2009},
  publisher={Stanford university}
}

@misc{cryptoeprint:2005/187,
      author = {Michael O.  Rabin},
      title = {How To Exchange Secrets with Oblivious Transfer},
      howpublished = {Cryptology ePrint Archive, Paper 2005/187},
      year = {2005},
      note = {\url{https://eprint.iacr.org/2005/187}},
      url = {https://eprint.iacr.org/2005/187}
}

\newpage
\appendix

\section{Appendix Overview}
This document supplements more details in the topics as follows:

\begin{enumerate}
    \item Appendix Overview
    \item Computing Infrastructure
    \item Pseudo Code
    \item Hyperparameters
    \item Additional Results
    \item Debugging Selective Approaches
\end{enumerate}

\section{Computing Infrastructure}
All experiments were executed on NVIDIA RTX A5000 graphics cards.

\section{Pseudo Code}
Algorithm~\ref{alg:our_alg_pseudo_code} is a detailed pseudo code of our scheme.
\begin{algorithm}
    \small
    \caption{\textsc{Block Coordinate Descent: Pseudo Code}}
    \begin{algorithmic}[1]
        \STATE\textbf{Inputs:} A network $f_{\theta, m}(\cdot)$ with parameters $\theta$ and a ReLUs mask $m$, \newline
        a train dataset $\mathcal{D}_{train}$,  \newline
        $B_\textrm{target}[\# \textrm{ReLUs}]$: target ReLU budget,   \newline
        $RT$ the number of random tries, \newline
        $DRC [\# \textrm{ReLUs}]$ is the Delta ReLU Count and \newline
        $ADT$ [\%] is the Accuracy Degradation Tolerance. 
        \STATE We start from a network with $||m||_0 = B_\textrm{ref}[\# \textrm{ReLUs}]$ ReLUs.
        \FOR{ $t = 1, ..., T = \lceil\frac{B_\textrm{ref} - B_\textrm{target}}{DRC}\rceil$}
            \STATE $n \leftarrow 0$
            \STATE $\textrm{found} \leftarrow \textrm{false}$
            \STATE Initialize a dictionary $d$ which maps binary ReLU maps to their accuracy drops: $d = \{\}$
           \WHILE {$n < \text{RT} \land \textrm{NOT}$$(\textrm{found})$}
                \STATE Randomly sample $DRC$ ReLUs from the pool of present ReLUs in the mask $m$. We define a mask $m^{\textrm{small}}_n$ such that $||m^{\textrm{small}}_n||_0 = DRC$.
                \STATE Set $m_{\textrm{hypothesis}} \leftarrow m \odot m^{\textrm{small}}_n$
                \STATE Evaluate the Accuracy degradation for the current ReLU hypothesis: $\Delta Acc \leftarrow \textrm{Accuracy}(f_{\theta, m}) - \textrm{Accuracy}(f_{\theta, m_{\textrm{hypothesis}}})$
                \IF {$\Delta Acc<ADT$}
                    \STATE found $\leftarrow \textrm{true}$
                    \STATE $m_{n}^{*} \leftarrow m^{\textrm{small}}_n$
                \ENDIF 
                \STATE Update $d[m^{small}_n] = \Delta Acc $
                \STATE $n \leftarrow n + 1$
           \ENDWHILE
           \IF {$\textrm{NOT}(\textrm{found})$}
                \STATE $m_{n}^{*} \leftarrow \argmin _{d.keys()} d.values()$
           \ENDIF 
           \STATE $m \leftarrow m \odot m_{n}^{*}$
           \STATE Fine-tune the network $f_{\theta, m}(\cdot)$
        \ENDFOR
    \end{algorithmic}
    \label{alg:our_alg_pseudo_code}
\end{algorithm}

\section{Hyperparameters}
\subsection{\textbf{$B_{\text{ref}}$ vs $B_{\text{target}}$} }
Table~\ref{tab:reference_budgets_resnet18} shows the $B_{\text{ref}}$ used to arrive at each $B_{\text{target}}$ in our ResNet18 experiments.
Table~\ref{tab:reference_budgets_wideresnet} shows the $B_{\text{ref}}$ used to arrive at $B_{\text{target}}$ in our WideResNet22-8 experiments.

\begin{table}[]
    \resizebox{0.95\linewidth}{!}{

\begin{tabular}{|l|l|l|}
\hline
\multicolumn{1}{|c|}{\textbf{Dataset}} & \multicolumn{1}{c|}{\textbf{Target Budget $B_\textrm{target}$}} & \multicolumn{1}{c|}{\textbf{Reference Budget $B_\textrm{ref}$}} \\ \hline
\multirow{2}{*}{CIFAR-10}               & $\geq$ 100K                        & 200K                                           \\ \cline{2-3} 
                                       & $<$ 30K                               & 30K                                            \\ \hline
\multirow{2}{*}{CIFAR-100}              & $\geq$ 100K                        & 200K                                           \\ \cline{2-3} 
                                       & $<$ 30K                               & 30K                                            \\ \hline
\multirow{4}{*}{TinyImageNet}          & 59.1K                                       & 80K                                            \\ \cline{2-3} 
                                       & 99.6K                                       & 150K                                           \\ \cline{2-3} 
                                       & 150K                                        & 180K                                           \\ \cline{2-3} 
                                       & 200K                                        & 220K                                           \\ \hline
\end{tabular}
}
\caption{\textbf{Reference budgets: } We summarize the respective $B_\textrm{ref}$ budgets for all our experiments for ResNet18. $B_\textrm{ref}$ is the ReLU budget we start from and $B_\textrm{target}$ is the target ReLU budget we wish to arrive at}

\label{tab:reference_budgets_resnet18}
\end{table}
\begin{table}[]
    \resizebox{0.95\linewidth}{!}{

\begin{tabular}{|l|l|l|}
\hline
\multicolumn{1}{|c|}{\textbf{Dataset}} & \multicolumn{1}{c|}{\textbf{Target budget $B_\textrm{target}$}} & \multicolumn{1}{c|}{\textbf{Reference budget $B_\textrm{ref}$}} \\ \hline
\multirow{2}{*}{CIFAR-10}               & 50K                                         & 75K                                            \\ \cline{2-3} 
                                       & $\geq$ 240K                        & 400K                                           \\ \hline
\multirow{2}{*}{CIFAR-100}              & 50K                                         & 75K                                            \\ \cline{2-3} 
                                       & $\geq$ 120K                        & 200K                                           \\ \hline
\multirow{2}{*}{TinyImageNet}          & 488.8K                                      & 570K                                           \\ \cline{2-3} 
                                       & $\geq$ 200K                        & 300K                                           \\ \hline
\end{tabular}
}
\caption{\textbf{Reference budgets: } We summarize the respective $B_\textrm{ref}$ budgets for all our experiments for WideResNet22-8. $B_\textrm{ref}$ is the ReLU budget we start from and $B_\textrm{target}$ is the target ReLU budget we wish to arrive at}
\label{tab:reference_budgets_wideresnet}

\end{table}

\subsection{Our method on top of SNL for WideResNet: Hyperparameters }
Table~\ref{tab:hyperparams_snl_wrn} shows the hyperparameters used for our SNL-based WideResNet22-8 experiments. In all of our experiments, for our finetune routine, we used ADAM~\cite{kingma2015adam} optimizer with an initial learning rate of $3.5\cdot 10^{-5}$ and a Cosine Annealing scheduler.

\begin{table}[]
    \resizebox{0.95\linewidth}{!}{

\begin{tabular}{|l|l|l|l|l|}
\hline
\multicolumn{1}{|c|}{\textbf{}} & \multicolumn{1}{c|}{\textbf{ADT {[}\%{]}}} & \multicolumn{1}{c|}{\textbf{DRC}} & \multicolumn{1}{c|}{\textbf{RT}} & \multicolumn{1}{c|}{\textbf{Finetune Epochs}} \\ \hline
CIFAR-10                        & 0.1                                        & 100                               & 50                               & 20                                            \\ \hline
CIFAR-100                       & 0.1                                        & 100                               & 50                               & 20                                            \\ \hline
TinyImageNet                    & 0.1                                        & 300                               & 50                               & 5                                             \\ \hline
\end{tabular}
}
\caption{\textbf{Hyperparameters for WideResNet22-8:} This table shows the hyperparameters used for executing the Coordinate Descent on top of SNL on WideResNet22-8. }

\label{tab:hyperparams_snl_wrn}
\end{table}

\subsection{Our method on top of AutoRep: Hyperparameters}
All of our AutoRep experiments (on both ResNet18 and WideResNet22-8) were done on CIFAR-100, and in that setting, we used a DRC of $100$, ADT of $0.1$, RT of $50$ and a $20$ finetune epochs. Similarly to the previous section, we used ADAM optimizer with an initial learning rate of $3.5\cdot 10^{-5}$ and a Cosine Annealing scheduler.

\section{Additional Results}
\begin{figure}
    \centering
    \begin{tikzpicture}
    \begin{semilogxaxis}[legend pos=south east,
        legend style={nodes={scale=0.62, transform shape}},
        width=0.7\columnwidth,
        height=5.7cm,
        xlabel= \small ReLU Count (K),
        ylabel=\small $\frac{\textrm{Accuracy for Budget}}{\textrm{Baseline Accuracy}}$ (\%),
        xlabel style={at={(0.5, 0.1)}},
        ylabel style={at={(-0.05, 0.5)}},
        title=WideResNet22-8 at CIFAR-100,
        xmin=32, xmax=350,
        ymin=90,ymax=100,
        ylabel near ticks,
        xlabel near ticks,
        axis background/.style={fill=blue!0},
        grid=both,
        log basis x = 2,
        /pgf/number format/1000 sep={\,},
        log ticks with fixed point,
        grid style={line width=.1pt, draw=gray!10},
        major grid style={line width=.2pt,draw=gray!50},
        ]

     \addplot[mark=*,mark size=2.5pt,blue, thick]
        plot coordinates {
            (50, 92.35)
            (120, 97.16)
            (150, 98.44)
            (180, 99.32)
            
        };
    \addlegendentry{\textbf{Ours}}

    \addplot[mark=x,mark size=2.5pt,orange, thick]
        plot coordinates {
            (180, 97.89)
            (240, 98.75)
            (300, 99.65)

        };
    \addlegendentry{SENet}

    \end{semilogxaxis}
\end{tikzpicture}
    \caption{ \textbf{Ours vs SENet on a WideResNet22-8 backbone:} We show that our
method outperforms SENet in a metric which is agnostic to the baseline classifier accuracy. Specifically, we measure the ratio between the performance reached by executing
each method and divide it by the accuracy of the baseline.}
    \label{fig:ours_vs_senet_wideresnet}
\end{figure}
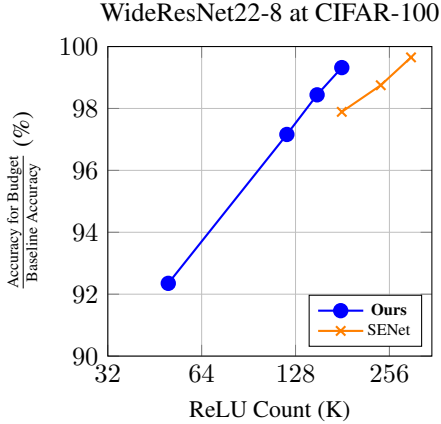

Since SENet~\cite{kundu2023senet} starts from a network of accuracy 80.82\% for CIFAR-100 on WideResNet22-8, and the standard training of networks in our method, and SNL's start at the 78\% ballpark, we take the course of a relative metric. Figure~\ref{fig:ours_vs_senet_wideresnet} shows our method's performance and SENEt. One can observe that our method achieves the same relative to baseline performance in lower budgets than SENet. Therefore, we conclude that our method outperforms SENet in this context.

\section{Debugging Selective Approaches}
\subsection{SNL:Lagrange Multiplier Finetuning}

\begin{figure}
    \centering
    \includegraphics[width=0.4\textwidth]{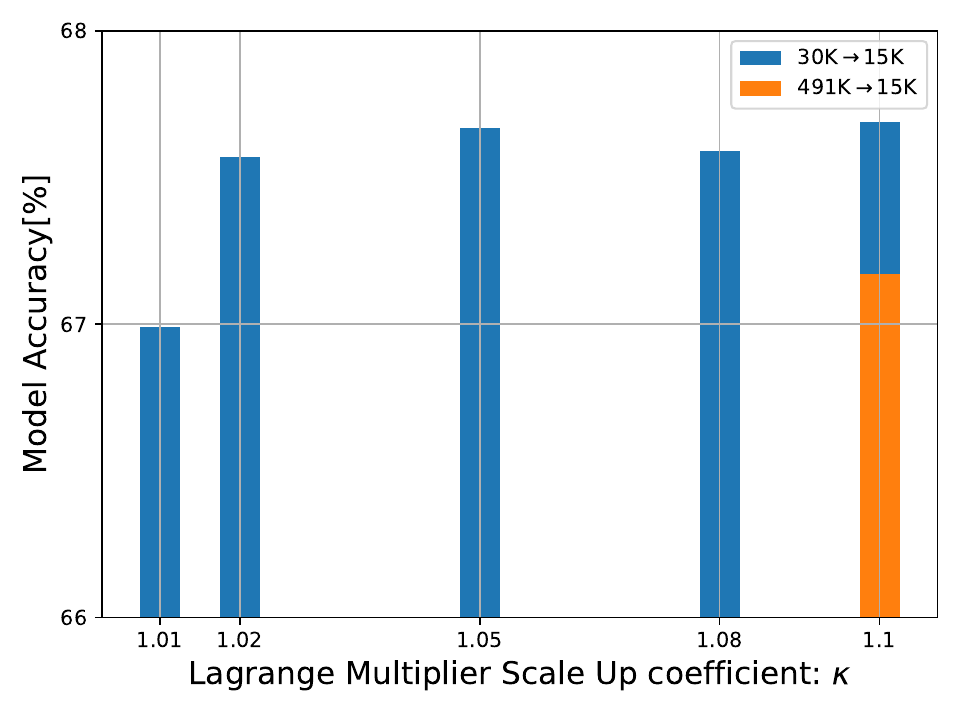}
    \caption{\textbf{Accuracy vs $\kappa$} This shows 2 types of SNL run configurations, starting from the original $490K$ ReLU network down to $15K$ and another starting from an SNL network which was optimized to $30K$ going down to $15K$.}
    \label{fig:kappa_snl}
\end{figure}

Our method is focused on the low budget regime. For very low budgets it is hard to set the balance between the requirement to meet a certain budget and network performance. We demonstrate that this is not straight forward through two experiments. SNL~\cite{cho2022selective} denote this factor as $\lambda$ which starts from some initial value $\lambda_0$. Then, when the number of ReLUs in each iteration of their algorithm is lower than some threshold, the value of $\lambda$ gets updated: $\lambda \leftarrow \kappa \cdot \lambda$. We scan multiple values of the correction to the hyperparameter $\kappa$ and summarize the effect of it on the network performance when the ReLU budget approaches $15K$ in Figure~\ref{fig:kappa_snl}. We observe that introducing a lower $\kappa$ to the lasso coefficient for lower budgets improves performance in about $0.5\%$. Our method achieves a boost of $2\%$ in the same setting - going from SNL $30K$ to $15K$.

But, how does one set the Lagrange multiplier value? This knob reflects the trade-off between the performance of the network (governed by the Cross-Entropy loss) and the optimization restriction which is to adhere to the target budget. When the budget has reached, this knob should be set to zero, reflecting that only the task performance should be optimized. Intuitively, we expect this to shrink to zero. We assume that the trend of the number of ReLUs \emph{reduced} in each iteration should be decreasing. 

\begin{figure}
    \centering
    \resizebox{0.98\linewidth}{!}{

    \begin{tabular}{c|c}
        ReLU Budget vs Epoch & $\Delta$ ReLU Budget vs Epoch \\
        \includegraphics[width=0.4\linewidth]{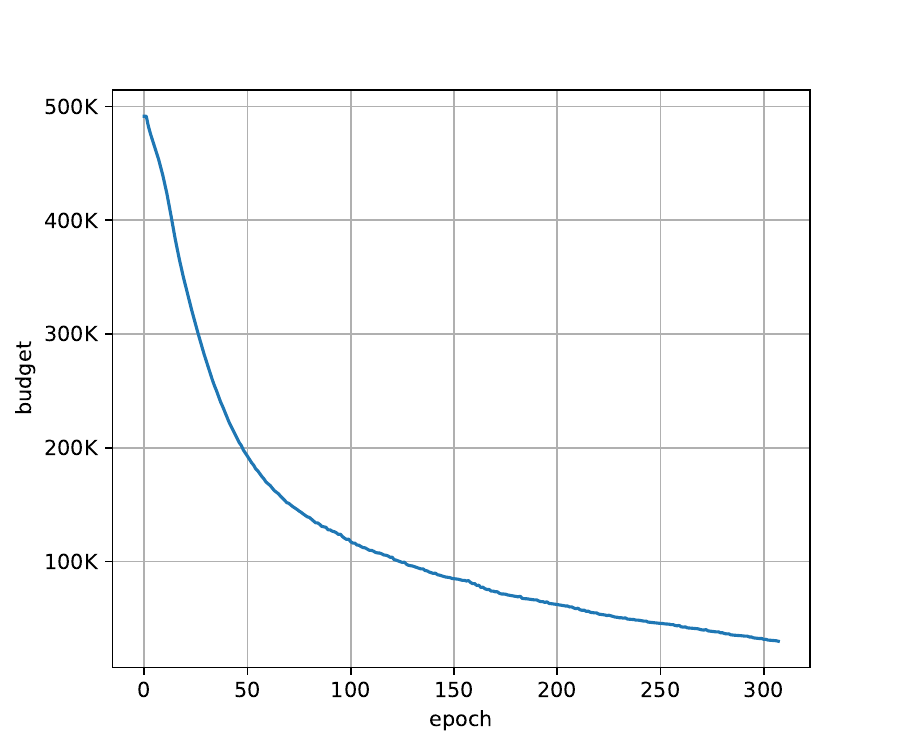} & \includegraphics[width=0.4\linewidth]{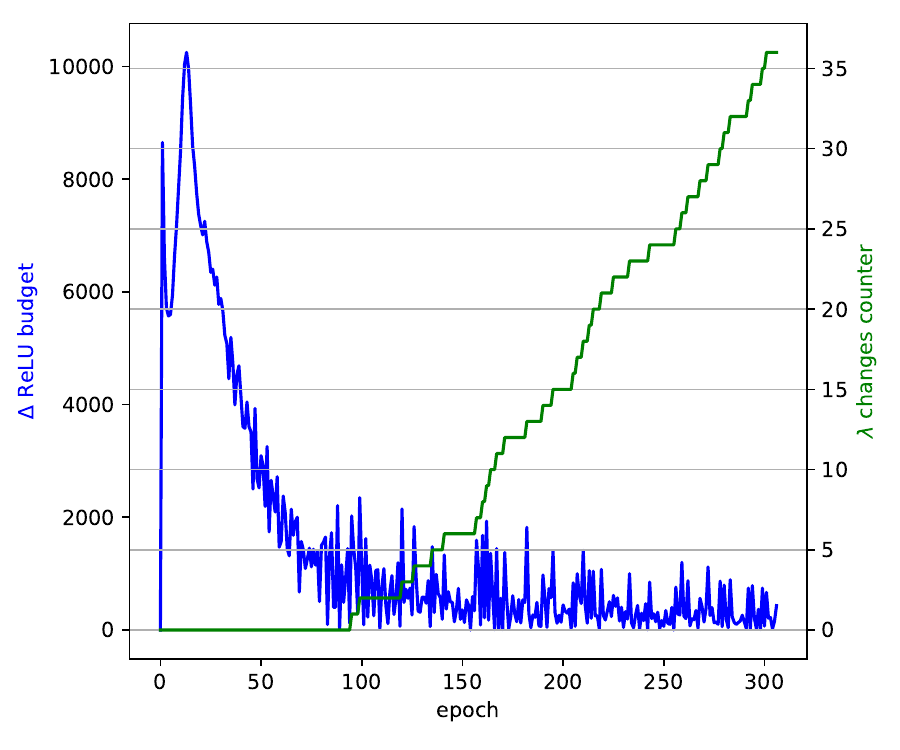} \\
        (a) &  (b)
    \end{tabular}
}    
    \caption{\textbf{ReLU and $\Delta\text{ReLU Budget}$ vs epoch:} (a) ReLUs Budget vs epoch, (b) blue: the number of ReLUs we decrease in each iteration, green: a counter which grows by $1$ whenever the Lagrange coefficient updates $\lambda\leftarrow \lambda \cdot \kappa$. }
    \label{fig:snl_relus_decrease_vs_epoch}
\end{figure}

Figure~\ref{fig:snl_relus_decrease_vs_epoch} shows an analysis of the ReLU budget and the ReLU budget decrease rate versus the epochs in SNL. Figure~\ref{fig:snl_relus_decrease_vs_epoch} (a) shows the ReLU budget per epoch. Figure~\ref{fig:snl_relus_decrease_vs_epoch} (b) shows the rate at which the ReLU budget decreases. As we anticipate, it starts to decrease monotonically. But, unfortunately, when the $\kappa$ mechanism turns on, this ReLU budget decrease in no longer monotonically decreasing and the ReLU decrease rate changes rapidly. This highlights how difficult it is to set the Lagrange multiplier and the effect it has on finding the real optimal solution for the ReLU mask configuration.

Our method, on the other hand, works directly on the ReLU mask and the rate in which we decrease ReLUs is a \emph{parameter} of our method. We acknoledge that a straightforward extension of our method would be to implement a scheduler for the ReLU decrease parameter. But, we demonstrate superior performance over selective approaches with a simple constant ReLU decrease of $100$ ReLUs in each iteration.

\subsection{What is the relationship between ReLU masks of Different Budgets?}

\begin{figure}
    \centering
    \includegraphics[width=0.9\linewidth]{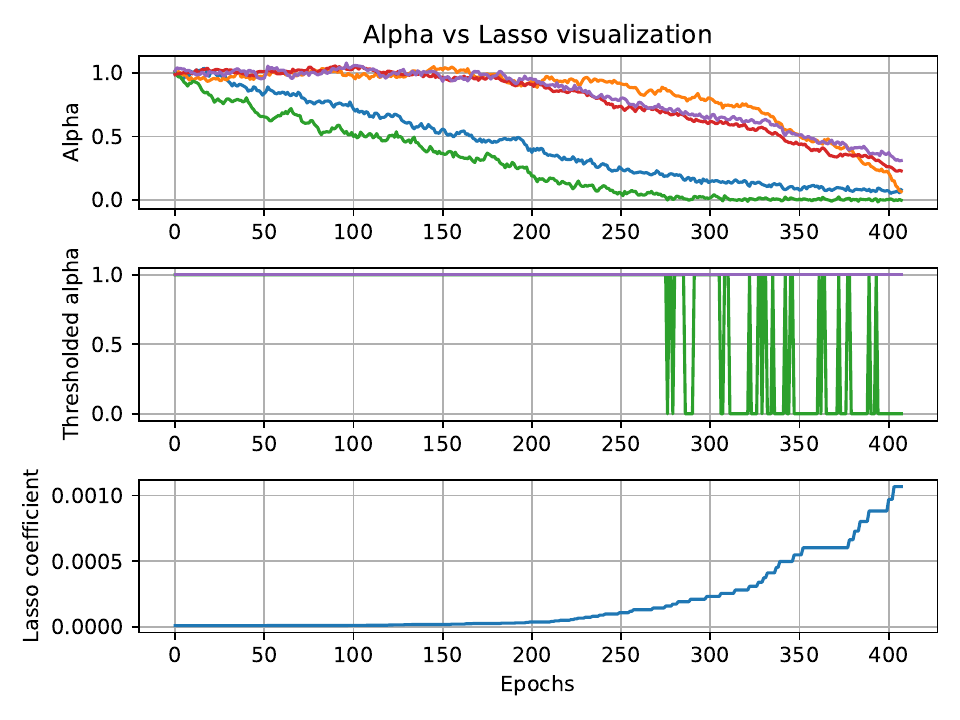}
    \caption{\textbf{Alphas vs. Lasso coefficient: } Alpha's are SNL's terminology for the ReLU masks. We pick some pixel locations for which ReLUs were kept by the SNL routine, and we can see that the alphas are gradually decreased. We also observe that some alphas hover around the threshold, which is correlated with the epochs for which the Lasso coefficient is updated. }
    \label{fig:alpha_vs_epoch}
\end{figure}

Our algorithm assumes that the overlap between the maps should be as large as possible. We illustrate the propagation over time of the soft-mask parameter values $\alpha$ of a Selective approach such as SNL in Figure~\ref{fig:alpha_vs_epoch}. In this figure, we observe a slow decay of the $\alpha$-values towards zero. We conjecture that some events of alphas crossing the threshold are associated to the updates of the Lasso coefficient using $\kappa$. Therefore, we suspect that such Selective methods yield sub-optimal solutions due to the re-calibration of the Lasso coefficient. Our method offers a direct approach which does not use the approximation of the $L_0$ norm with a lasso regularization. Whereas for SNL the binary masks decay slowly, in our case we work directly in the binarized alphas domain so we decide which ReLUs are eliminated \emph{faster}. In the Discussion section we provide a theoretical analysis of the convergence rate of our method.

\end{document}